 \documentclass[pmlr,twocolumn,10pt]{jmlr} 

\usepackage{booktabs}
\usepackage{siunitx}

\usepackage[switch]{lineno}

\usepackage{float}
\usepackage{caption}
\usepackage{enumitem}       

\usepackage{multirow}

\usepackage{xcolor,colortbl}
\usepackage{color, colortbl} 
\usepackage{nicematrix}

\usepackage{xcolor} 

\usepackage[most]{tcolorbox} 
\usepackage{enumitem}

\usepackage{subcaption}

\usepackage{makecell}

\newcommand{\Sref}[1]{\S\ref{#1}}

\definecolor{highlight}{rgb}{1.0,0.90,0.8}	
\definecolor{Gray}{gray}{0.96}

\definecolor{lightBlue}{rgb}{0.78, 0.85, 1.0}
\definecolor{lightOrange}{rgb}{0.88, 0.95, 1.0}
\definecolor{lightRed}{rgb}{1.0, 0.85, 0.85}
\usepackage{tcolorbox}
\usepackage{multirow}
\usepackage{array}

\newcommand{\parahead}[1]{\medskip\noindent\textbf{#1}\par\smallskip}

\newtcbox{\bluebox}{on line, box align=base, colback=lightBlue,colframe=white,size=fbox,arc=3pt, before upper=\strut, top=-2pt, bottom=-4pt, left=-2pt, right=-2pt, boxrule=0pt}

\newtcbox{\orangebox}{on line, box align=base, colback=lightOrange,colframe=white,size=fbox,arc=3pt, before upper=\strut, top=-2pt, bottom=-4pt, left=-2pt, right=-2pt, boxrule=0pt}

\newtcbox{\redbox}{on line, box align=base, colback=lightRed,colframe=white,size=fbox,arc=3pt, before upper=\strut, top=-2pt, bottom=-4pt, left=-2pt, right=-2pt, boxrule=0pt}

\newtcbox{\whitebox}{on line, box align=base, colback=white,colframe=white,size=fbox,arc=3pt, before upper=\strut, top=-2pt, bottom=-4pt, left=-2pt, right=-2pt, boxrule=0pt}

\newcommand{\dashifted}{\raisebox{0.5\depth}{\tiny$\downarrow$}}
\newcommand{\upshifted}{\raisebox{0.5\depth}{\tiny$\uparrow$}}

\newcommand{\dab}[1]{{\scriptsize\bluebox{\dashifted{#1}}}}

\newcommand{\dar}[1]{{\scriptsize\redbox{\dashifted{#1}}}}

\newcommand{\uab}[1]{{\scriptsize\bluebox{\upshifted{#1}}}}

\newcommand{\equal}[1]{{\hypersetup{linkcolor=black}\thanks{#1}}}

\theorembodyfont{\upshape}
\theoremheaderfont{\scshape}
\theorempostheader{:}
\theoremsep{\newline}

\jmlrvolume{297}
\jmlryear{2025}
\jmlrworkshop{Machine Learning for Health (ML4H) 2025} 

 \title[Automated Natural Language Surgical Feedback Generation]{Generating Natural-Language Surgical Feedback: \\From Structured Representation to Domain-Grounded Evaluation}

\author{
    \Name{Firdavs Nasriddinov}\equal{These authors contributed equally} \Email{firdavs@caltech.edu}\\
    \addr California Institute of Technology, USA
\AND
    \Name{Rafal Kocielnik}\footnotemark[1]  \Email{rafalko@caltech.edu, Rafal.Kocielnik@cshs.org}\\
    \addr California Institute of Technology, Cedars-Sinai Medical Center, USA
\AND
    \Name{Anima Anandkumar} \Email{anima@caltech.edu}\\
    \addr California Institute of Technology, USA
\AND
    \Name{Andrew J. Hung} \Email{andrew.hung@cshs.org}\\
    \addr Cedars-Sinai Medical Center, USA
    }

\begin{document}

\maketitle

\begin{abstract}
High-quality intraoperative feedback from a surgical trainer is pivotal for improving trainee performance and long-term skill acquisition. Automating natural, trainer-style feedback promises timely, accessible, and consistent guidance at scale—but requires models that understand 
clinically relevant representations. We present a structure-aware pipeline that learns a surgical action ontology from real trainer$\rightarrow$trainee transcripts (33 surgeries) and uses it to condition feedback generation. We contribute by 1) mining Instrument-Action-Target (IAT) triplets from real-world feedback text and clustering surface forms into normalized categories, 2) fine-tuning a video$\rightarrow$IAT model that leverages the surgical procedure and task contexts, as well as fine-grained temporal instrument motion (crucial for representing instruments and actions over time), and 3) demonstrating how to effectively leverage IAT triplet representation to guide GPT-4o in generating clinically-grounded natural, trainer-style feedback. We show that, on \textit{Task 1: Video$\rightarrow$IAT recognition}, our context injection and temporal tracking deliver consistent AUC gains -- Instrument: 0.67$\rightarrow$0.74, Action: 0.60$\rightarrow$0.63, Tissue: 0.74$\rightarrow$0.79. For \textit{Task 2: Feedback text generation}  (1 [opposite/unsafe] - 3 [admissible] - 5 [perfect match] fidelity rubric against human trainer), GPT-4o from video alone scores 2.17; IAT conditioning reaches 2.44 (+12.4\%), increasing the admissible generations with score $\geq$3: 21\%$\rightarrow$42\%. Traditional metrics also improve: Word Error Rate (WER): $\downarrow$15–31\% and ROUGE (phrase/substring overlap): $\uparrow$9–64\%. Grounding generation in explicit IAT structure improves fidelity and yields clinician-verifiable rationales, supporting auditable use in surgical training.
\end{abstract}


\paragraph*{Data and Code Availability}
Due to privacy restrictions, data is available upon request. The code is publically available on \href{https://github.com/firdavsn/SurgFBGen}{GitHub}.

\paragraph*{Institutional Review Board (IRB)}
The data used was collected under IRB of the University of Southern California (HS-17-00113).

\vspace{-8pt}
\section{Introduction}
\vspace{-4pt}
\label{sec:intro}

\begin{figure*}[ht!]
\centering
\includegraphics[width=\textwidth]{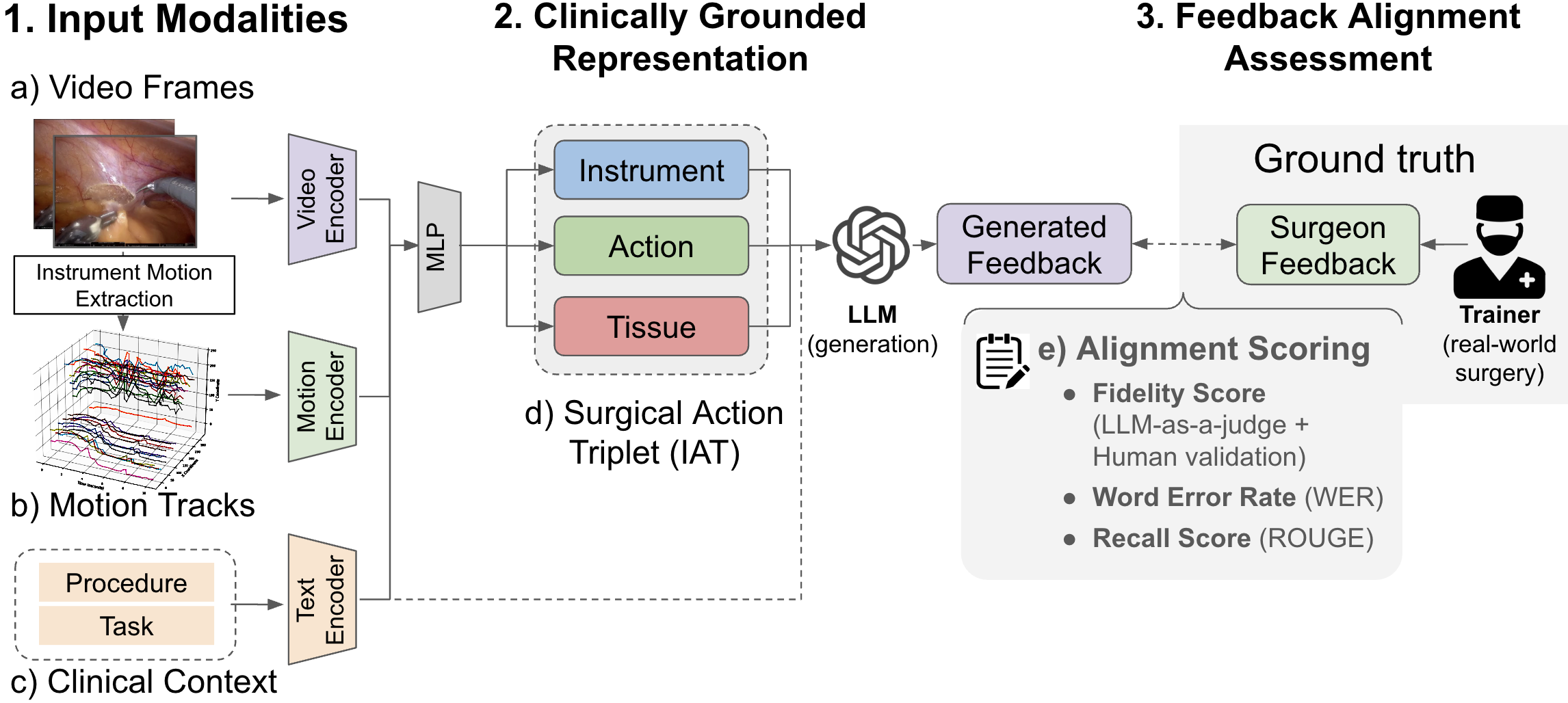}
\vspace{-16.0pt}
\caption{\textbf{Structure-aware pipeline for clinically aligned surgical feedback.}
\textbf{(1) Multimodal inputs:} video, inferred instrument motion, and procedure/task context are encoded and fused. 
\textbf{(2) Clinically grounded representation:} the fused features yield an \emph{Instrument–Action–Tissue (IAT)} triplet summarizing the tool–tissue interaction.
\textbf{(3) Feedback generation \& evaluation:} predicted IATs condition an LLM to produce trainer-style feedback, assessed with a clinician-aligned \emph{fidelity} and standard text metrics.}
\label{fig:work_overview}
\vspace{-10.0pt}
\end{figure*}

\paragraph{Importance:}

High-quality intraoperative feedback from the surgical trainer is pivotal for improving trainee performance and long-term skill acquisition. Real-time trainer feedback aimed at modifying trainee behavior helps avoid errors \citep{wong2023development}, and the quality of feedback is linked to intraoperative performance \citep{bonrath2015comprehensive}, long-term learning outcomes \citep{agha2015role}, and operative outcomes via communication quality \citep{d2020evaluating}. Automatically generation natural, trainer-style feedback promises timely, accessible, and consistent guidance at scale—extending expert coaching beyond what busy trainers can provide in real time—while mitigating the inefficiency and rater biases inherent to manual assessment \citep{chinh2019ways}. Realizing this promise requires approaches that leverage clinically meaningful representations, so that generated guidance remains clinically grounded, interpretable, and auditable.

\paragraph{Challenges:}
Surgical feedback is high-stakes: guidance must be technically correct and grounded in the live operative scene. In practice, feedback is spontaneous, conversational speech that entangles clinically critical \emph{semantics}—which instrument, which action, which target—with surface \emph{syntax} (style, tone, hedges) \citep{wong2023development}, while the scene itself is visually complex and evolves through tightly coupled instrument–tissue interactions \citep{ma2022surgical}. Large, diverse corpora of synchronized surgical video and trainer feedback are rare due to privacy/IRB constraints and the cost of expert annotation, yielding \emph{data scarcity} and amplifying \emph{domain shift} between general purpose and surgery-specific video-language representations \cite{philipp2022dynamic}. Off-the-shelf VLMs often miss procedure/task context and lack mechanisms to bind utterances to concrete visual evidence under these shifts \citep{kiyasseh2023vision}. Moreover, text-overlap metrics (e.g., WER/ROUGE) poorly reflect clinical correctness, motivating structure-aware, clinically grounded representations (e.g., Instrument–Action–Tissue semantics) and clinician-aligned evaluation.

\paragraph{Our Approach:}

We introduce a \emph{structure-aware pipeline}
\footnote{'structure-aware' - using explicit, clinically meaningful intermediate schema to guide perception and generation.} that (i) learns a feedback-induced surgical action ontology from real trainer\,$\to$\,trainee transcripts and (ii) uses that structure to guide natural language feedback generation from surgical videos. 

An overview of our feedback generation pipeline is depicted in Fig. \ref{fig:work_overview}. We first train a video$\rightarrow$IAT predictor that fuses three inputs: (1a) video frames from endoscopic camera (mean-pooled embeddings), (1b) \emph{instrument motion tracks} derived by estimating depth and tracking multi-point trajectories (details in Fig. \ref{fig:motion_tracking}), and (1c) \emph{clinical context} (procedure and local task) as text embeddings. These streams are concatenated with an MLP to produce multi-head predictions of the (2d) \emph{Instrument–Action–Tissue (IAT)} triplet. The predicted triplets then condition GPT-4o to generate natural language feedback, with an uncertainty gate forwarding only confident triplets to mitigate hallucinations; outputs are compared to trainer ground truth using a clinician-aligned 1-–5 \emph{fidelity} score and secondary WER/ROUGE (3e).

To enable this pipeline, we automatically mine \emph{Instrument–Action–Target (IAT)} mentions from trainer\,$\to$\,trainee teaching transcripts delivered in 33 real surgeries collected by \cite{wong2023development}. We cluster various mentions into normalized categories, yielding a data-driven dictionary aligned with clinical language which supports Video$\rightarrow$IAT mapping.

\paragraph{Findings:}
\begin{itemize}[leftmargin=*, itemsep=-0.5mm, topsep=2pt]
    \item \textbf{IAT triplet recognition from video:} Adding temporal instrument tracking and surgical context yields consistent gains across backbones (Table \ref{tab:video_iat_auc_gain_results})—roughly \emph{+0.07–0.08 AUC} for Instrument (\(\sim\!+10\)–\(12\%\)), \emph{+0.02–0.03} for Action (\(\sim\!+4\)–\(5\%\)), and \emph{+0.05–0.09} for Tissue (\(\sim\!+7\)–\(13\%\)).
    
    \item \textbf{Feedback generation:} Structure-aware prompting raises scores on 1-–5 fidelity rubric (Table \ref{tab:alignment_text_scores}) from \(\mathit{2.17}\) (video-only) to \(\mathit{2.42}\) with high-confidence triplets (\(+12.4\%\)), increasing the share of admissible ($\ge\!3$: \(+22\) pp) and high-quality feedback ($\ge\!4$: \(+3\) pp). Secondary metrics also improve (WER \(\downarrow\) \(31\%\); ROUGE \(\uparrow\) \(64\%\)).
\end{itemize}

\paragraph{Contributions:}
\begin{itemize}[leftmargin=*, itemsep=-0.5em, topsep=2pt]
    \item \textbf{Surgeon-aligned feedback from video:} To the best of our knowledge, we are the first to propose an end-to-end generation of natural-language feedback in robot-assisted surgery, grounded in interpretable IAT triplets and motion cues.
    \item \textbf{Two-step, structure-aware framework:} (1) video, procedure/task text, instrument motion$\rightarrow$IAT prediction; (2) controlled LLM feedback generation conditioned on triplets.
    \item \textbf{Clinically grounded evaluation:} A reproducible protocol combining a clinician-aligned 1–-5 fidelity rubric with an interpretable, surgery-specific LLM-as-judge alignment score.
\end{itemize}

\vspace{-8pt}
\section{Related Work}
\vspace{-4pt}
\label{sec:related_work}

\paragraph{Feedback in Robot-Assisted Surgery.}
Natural-language feedback is the primary medium of trainer guidance in surgery \citep{wong2023development}, and its quality is linked to intraoperative performance \citep{bonrath2015comprehensive} and operative outcomes \citep{d2020evaluating}. Prior work has largely \emph{measured} rather than \emph{delivered} feedback: content categorization \citep{wong2023development,kocielnik2023deep}, delivery-style quantification \citep{knudsen2025mp06,kocielnik2024human}, and multimodal prediction of effectiveness \citep{gupta2025multi}. On the delivery side, rubric-based tools (e.g., EASE) support targeted coaching \citep{haque2022assessment}; wizard-of-oz studies show real-time standardized coaching can help \citep{laca2022using}; early automated systems provide narrow, clip-level video feedback \citep{ma2024artificial}; and complementary hardware work explores force feedback \citep{servais2025novel}. We instead generate \emph{surgeon-aligned natural-language feedback} directly from operative video by grounding an LLM in clinically meaningful \emph{Instrument–Action–Target} structure predicted via multimodal fusion.

\paragraph{Vision--Language Models in Surgery.}
Recent surgical VLMs target different goals. \textit{GP-VLS} trains on six surgery-focused datasets and evaluates with \textit{SurgiQual}, showing strong knowledge and vision–language performance \citep{schmidgall2024gp}. \textit{SurgRAW} uses a multi-agent, chain-of-thought framework with retrieval to reduce hallucinations across workflow tasks such as instrument/action recognition and outcome assessment \citep{low2025surgraw}. \textit{Surgical-LVLM} adapts a large VLM with VP-LoRA and a token-interaction module to couple answers with localized evidence, achieving SOTA on EndoVis VQA/VQLA benchmarks \citep{wang2024surgical}. Earlier systems (\textit{SurgicalGPT} \citep{seenivasan2023surgicalgpt}, \textit{Surgical-VQA} \citep{seenivasan2022surgical}, \textit{Surgical-VQLA} \citep{bai2023surgical}) often reframe classification datasets or attach heads, yielding descriptive scene understanding rather than prescriptive guidance. In parallel, foundational surgical video models improve representation transfer: \textit{GSViT} pre-trains on 680h with next-frame objectives \citep{schmidgall2024general}; \textit{SurgVLP} aligns 3k hours of lecture video with transcripts \citep{yuan2023learning}; and \textit{PeskaVLP} adds hierarchical knowledge augmentation with procedure-aware, DTW-aligned pretraining \citep{yuan2024procedure}. For feedback, prior video-based work stops at \emph{classification} of effectiveness \citep{kocielnik2023deep, wong2023development}. In contrast, we generate \emph{surgeon-aligned} natural-language feedback from operative video by grounding an LLM in predicted \emph{Instrument–Action–Target} structure and evaluate with a clinician-aligned fidelity rubric.

\paragraph{Surgical Action Triplets.}
Surgical action triplets—〈instrument, action/verb, target〉—are a standard fine-grained representation of tool–tissue interactions in endoscopic/robotic surgery, offering richer context than phase/step labels \citep{nwoye2020recognition}. The community formalized this via CholecTriplet 2021/2022, progressing from recognition (per-frame triplet) to detection (localizing tools and associating verb–target) \citep{nwoye2023cholectriplet2021, nwoye2023cholectriplet2022}. Advances include attention-based association \citep{nwoye2022rendezvous}, mixed-supervised instrument–tissue detection \citep{sharma2023surgical}, and multi-task spatiotemporal models (e.g., MT-FiST) \citep{li2023mt}. Prior work largely stops at triplet prediction from video; in contrast, we (i) \emph{induce a triplet ontology from trainer–trainee transcripts}, (ii) \emph{predict triplets via multimodal fusion}, and (iii) \emph{use triplets to control LLM feedback generation}—turning triplets from an analysis endpoint into a grounding interface for safe, surgeon-aligned natural-language feedback.

\vspace{-8pt}
\section{Methods}
\vspace{-4pt}
\label{sec:methods}

\begin{table}[t]
\centering
\small

\caption{Summary statistics of our dataset with breakdowns of counts of IAT triplets extracted.}
\vspace{-6.0pt}

\label{tab:dataset_summary}

\setlength{\tabcolsep}{9pt}
\renewcommand{\arraystretch}{1.2}
\begin{tabular}{l r r}
\toprule
\textbf{Category} & \textbf{Count} & \textbf{\% Total} \\
\midrule
Lines of Feedback Text            & 4210     & 100.0\% \\
Unique Procedures        & 7        & -- \\
Unique Tasks             & 36       & -- \\
\midrule
Unique Instruments\textsuperscript{*} & 114 (7)   & -- \\
Unique Actions\textsuperscript{*}     & 1107 (21)  & -- \\
Unique Targets\textsuperscript{*}     & 251 (11)  & -- \\
\midrule
Feedback w/ Instrument       & 344      & 8.2\% \\
Feedback w/ Action           & 875      & 20.8\% \\
Feedback w/ Tissue    & 512      & 12.2\% \\
\midrule
Single value triplet           & 1405     & 33.4\% \\
Multiple value triplets        & 416      & 9.9\% \\
\bottomrule
\end{tabular}

\raggedright
{\footnotesize \textsuperscript{*} Number before parenthesis represents raw counts. Number in parenthesis represents counts after clustering and selection.}
\vspace{-14.0pt}
\end{table}

\subsection{Data Acquisition}
\vspace{-4pt}
\label{sec:data_acq}
We use authentic trainer$\to$trainee feedback from real robot-assisted surgeries collected by \cite{wong2023development} (Table~\ref{tab:dataset_summary}). Audio was recorded via surgeon-worn wireless microphones and synchronized with endoscopic point-of-view video using an external device integrated with the da Vinci~Xi system \citep{dimaio2011vinci}. In total, 4{,}210 feedback instances were transcribed and paired with procedure labels (e.g., \textit{radical prostatectomy}, \textit{nephrectomy}) and coarse task annotations (e.g., \textit{adenoma dissection}, \textit{closing peritoneum}).

\begin{figure*}[ht!]
\centering
\includegraphics[width=\textwidth]{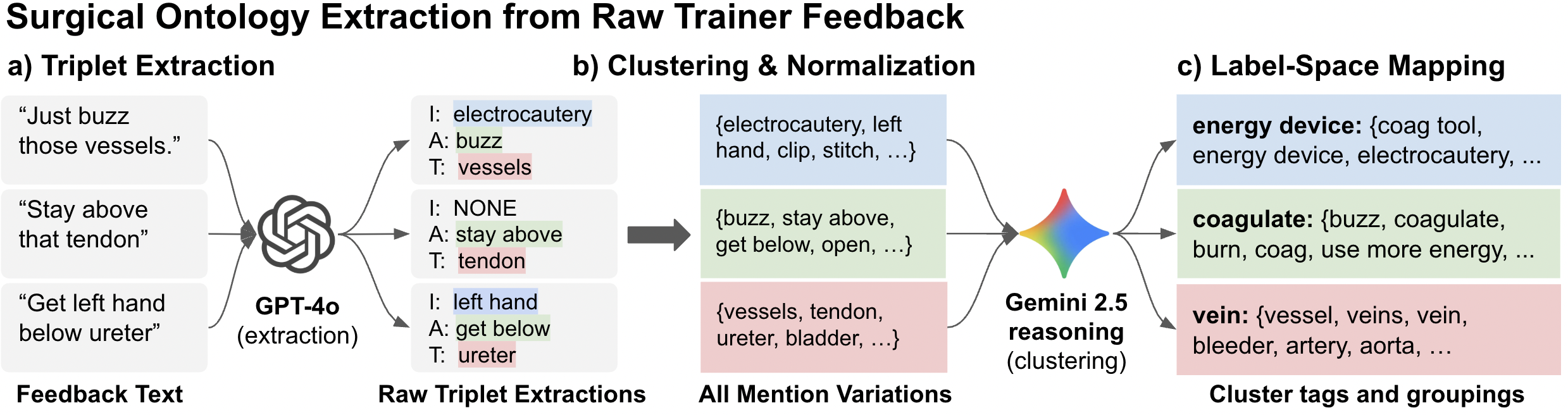}
\vspace{-16pt}
\caption{\textbf{Surgical ontology extraction from raw trainer feedback.}
\textbf{(a) Triplet Extraction} — GPT\mbox{-}4o parses free-text feedback into one or more \emph{Instrument–Action–Tissue} triplets \([\mathrm{I},\mathrm{A},\mathrm{T}]\), permitting \emph{null} components when unmentioned.
\textbf{(b) Clustering \& Normalization} — a reasoning LLM (Gemini~2.5) clusters semantically related surface forms for each slot (I/A/T), merges them into functionally coherent meta-clusters, and prunes low-frequency categories.
\textbf{(c) Label-Space Mapping} — canonical tags (e.g., \texttt{energy device}, \texttt{coagulate}, \texttt{vein}) and raw$\rightarrow$tag mappings are produced; this normalized label space provides weak supervision for the video$\rightarrow$\,IAT model and conditions feedback generation.}
\label{fig:ontology_extraction}
\vspace{-10pt}
\end{figure*}

\vspace{-8pt}
\subsection{Surgical Feedback Definition}
\label{sec:fbk_def}
We adopt the clinically validated definition of \citet{wong2023development}: surgical feedback is \emph{any dialogue intended to modify a trainee’s thinking or behavior during live surgery}. In our dataset, this comprises real-time trainer$\to$trainee utterances directed to a console-operating trainee and \emph{contextually tied} to the ongoing task to guide actions, decisions, or understanding. Social/irrelevant conversation is excluded.

\vspace{-8pt}
\subsection{Natural Language Surgical Feedback Generation Pipeline}
\label{sec:pipeline}
As shown in Fig.~\ref{fig:work_overview}, our pipeline (i) fuses multimodal inputs (video, motion, clinical context), (ii) grounds predictions in an interpretable surgical IAT representation, and (iii) evaluates generated trainer-style feedback against trainer references.

\paragraph{(1) Input Modalities.} 
We predict the \emph{Instrument–Action–Tissue (IAT)} triplet from short video clips (10\,s at 5\,fps) by fusing three complementary sources: endoscopic video frames, temporal instrument motion, and clinical context of procedure and task. Next, we describe how each is obtained.

\textbf{(a) Video Encoding.}
\label{subsec:video_context}
Frames are passed through a surgically pretrained backbone—\emph{PeskaVLP} \citep{yuan2024procedure} or \emph{SurgVLP} \citep{yuan2025learning}—to obtain per-frame embeddings, which we mean-pool across time into a clip-level \emph{video embedding}. We leverage surgical backbones because general-purpose models exhibit reduced sensitivity to clinically salient events across clips \citep{philipp2022dynamic}.

\textbf{(b) Motion Tracks.}
From the same clip we extract multi-point trajectories of instrument motion and embed them with a dedicated temporal motion encoder (LSTM) to represent tool kinematics. \emph{\textbf{Integrating temporal motion tracking is one of our key contributions}} (details in Sec. \Sref{subsec:temporal_tracking}). This step produces \emph{motion embeddings}.

\textbf{(c) Clinical Context Encoding.}
\label{subsec:clinical_context}
For each clip, we expand the provided \emph{procedure} and \emph{local task} into concise, clinically phrased summaries with GPT\mbox{-}4o (definitions in Tables~\ref{tab:proc_definitions}, \ref{tab:task_definitions_1}--\ref{tab:task_definitions_4}; prompts in Tables~\ref{tab:procedure_definition_prompt}, \ref{tab:task_definition_prompt}), following the knowledge-augmentation strategy of \citet{yuan2024procedure}. We then encode these summaries using the \emph{text encoder of the same surgically pretrained video–language backbone} as the video stream (i.e., \emph{SurgVLP} or \emph{PeskaVLP}) to preserve an aligned vision–text latent space and leverage surgical priors, yielding \emph{procedure} and \emph{task} embeddings.

\paragraph{(2) Clinically Grounded Representation (IAT triplet).}

We fuse the clip-level \emph{video}, \emph{procedure/task}, and \emph{motion} embeddings and pass the result to three identical MLP heads for I/A/T (3 layers: $64\!\rightarrow\!32\!\rightarrow\!16$). The heads produce the \textbf{(d) predicted Instrument–Action–Tissue} triplet—an interpretable abstraction of \textit{``which instrument did what to which tissue''} that grounds downstream feedback generation. Supervision derives from ground-truth triplets mined from trainer feedback via \emph{\textbf{our contribution in the form of ontology extraction pipeline}} (Sec.~\Sref{sec:ontology_from_raw_feedback}).

\paragraph{(3) Feedback Generation and Alignment Assessment.}
We cast feedback generation as \emph{data-to-text}: GPT\mbox{-}4o is conditioned on (i) \emph{procedure/task} summaries situating the clip in workflow, (ii) predicted \emph{IAT} triplets, (iii) \emph{video frames} from a 10\,s window at 1\,fps, and (iv) a \emph{reference lexicon} plus a few \emph{IAT$\rightarrow$feedback} examples. A single prompt (App.~\ref{app:gpt4o_fbk_gen_prompt}) frames the model as a surgical training assistant, constrains outputs to 1–3 sentences, and steers toward actionable, safety-oriented guidance tied to the observed tool–tissue interaction while avoiding redundancy. Generated feedback is then compared (via ~\ref{tab:feedback_eval_prompt}) to trainer references using a clinician-aligned \emph{fidelity} score and standard NLP metrics (Sec.~\Sref{sec:task_2_setup}).

\vspace{-4pt}
\subsection{Ontology Extraction and Weakly Supervised Label Construction}
\vspace{-2pt}
\label{sec:ontology_from_raw_feedback}

Figure~\ref{fig:ontology_extraction} overviews our \emph{\textbf{novel three-step ontology extraction pipeline}}, which provides a weak supervision labels for Video$\rightarrow$IAT prediction task.

\begin{figure*}[ht!]
\centering
\includegraphics[width=\textwidth]{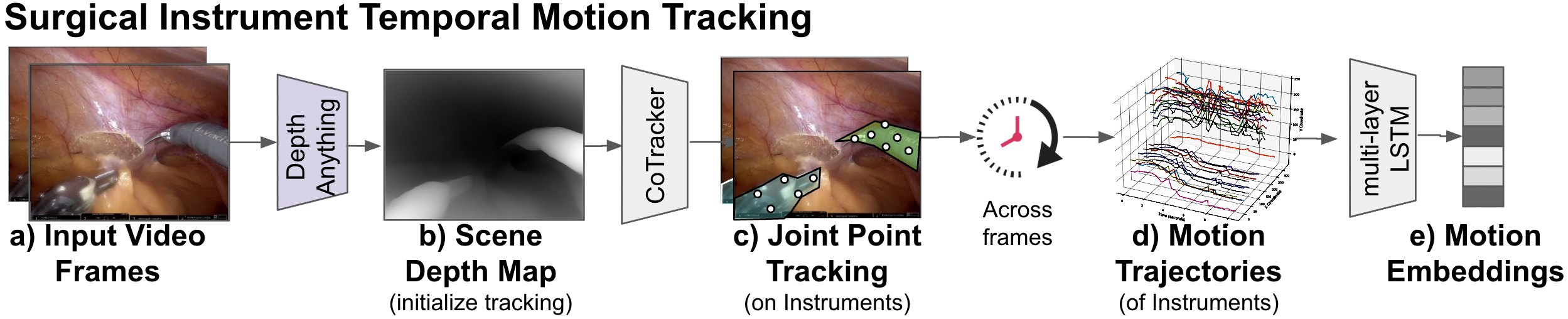}
\vspace{-20pt}
\caption{\textbf{Surgical instrument temporal motion tracking pipeline.}
\textbf{(a)} \emph{Input video frames} from the endoscopic view are processed to capture tool appearance and motion.
\textbf{(b)} A \emph{scene depth map} is estimated (Depth Anything) to initialize spatially consistent point tracking.
\textbf{(c)} \emph{Joint point tracking} of instrument keypoints is performed across frames using CoTracker. 
\textbf{(d)} The resulting \emph{motion trajectories} capture fine-grained temporal dynamics of instrument movement. 
\textbf{(e)} An LSTM encodes these trajectories into compact \emph{motion embeddings} for downstream modeling.}
\label{fig:motion_tracking}
\vspace{-10pt}
\end{figure*}

\begin{itemize}[leftmargin=*, itemsep=0.0em, topsep=4pt]

    \item \textbf{Triplet Extraction (Fig \ref{fig:ontology_extraction}.a)} We treat live trainer feedback transcripts as a source of \emph{weak supervision} for surgical action recognition. Each feedback snippet is processed with GPT\mbox{-}4o to extract one or more \emph{Instrument–Action–Target (IAT)} triplets in the form \texttt{[I, A, T]} (prompts in Tables ~\ref{tab:instrument_extraction_prompt}, ~\ref{tab:action_extraction_prompt}, ~\ref{tab:tissue_extraction_prompt}). This converts naturalistic comments into \emph{sparse multi-class labels} suitable for video classification; e.g., \textit{“use your grasper to gently pull back on the peritoneum”} $\rightarrow$ (\textit{grasper}, \textit{gently pull back}, \textit{peritoneum}). As feedback is not designed to enumerate all elements, triplets may be \emph{incomplete}; we permit \emph{null} fields for missing components.

    \item \textbf{Clustering \& Normalization (Fig \ref{fig:ontology_extraction}.b)} The raw extractions exhibit substantial lexical variation (e.g., \emph{“buzz”/“burn”/“coagulate”} and long-tail distributions, requiring both grouping of semantically related mentions and normalization of exact synonyms. We use Gemini 2.5 Pro~\citep{comanici2025gemini} for clustering and ontology construction, inspired by ~\cite{babaei2024llms4om, yang2024neural}. We introduce a semi-automated two-step procedure. First, the LLM clusters surface forms into fine-grained, semantically coherent groups (prompt in Table ~\ref{tab:initial_clustering_prompt}). Second, with few-shot prompting (Table ~\ref{tab:merge_clustering_prompt}), it merges these into relevant meta-clusters, followed by elbow-based pruning to drop low-frequency categories (Figure ~\ref{fig:iat_class_counts_with_thresholds}). Cluster coherence comparison against other methods can be found in Table~\ref{tab:clustering_coherence}, App.~\ref{app:ontology_extraction_details}.

    \item \textbf{Label-space Mapping (Fig \ref{fig:ontology_extraction}.c)} The resulting hierarchy maps varied surface forms to canonical, clinically meaningful tags (see App. \ref{app:ontology_tag_to_mention}), balancing fidelity to nuance (fine-grained clusters) against label frequency needed for robust video classification (meta-clusters). We preserve links across raw feedback text and canonical tags—yielding a normalized label space for video$\rightarrow$IAT supervision and  conditioning of feedback generation.
\end{itemize}

\vspace{-10pt}
\subsection{Temporal Instrument Motion Tracking}
\label{subsec:temporal_tracking}
 
Figure~\ref{fig:motion_tracking} illustrates our \textbf{\emph{novel pipeline for extracting instrument motion trajectories}} from surgical video. While video and context embeddings capture visual and procedure priors, they underutilize the rich kinematics of surgical tools. We contribute with a dedicated \emph{motion stream} in five sequential stages:

\begin{itemize}[leftmargin=*, itemsep=0.0em, topsep=4pt]
    \item \textbf{Input Frames (Fig.~\ref{fig:motion_tracking}.a)} Video clips (10s at 5 fps) provide the raw input for motion extraction.
    \item \textbf{Scene Depth Map (Fig.~\ref{fig:motion_tracking}.b)} We pre-emphasize metallic tools by dimming the first frame (distance-to-gray heuristic) and then run DepthAnything \citep{yang2024depth} to produce a dense depth map, providing geometric grounding and more reliable instrument point initialization.
    
    \item \textbf{Joint Point Tracking (Fig.~\ref{fig:motion_tracking}.c)} From the depth map, we construct an \emph{instrument-edge mask} via OpenCV edge detection with circular contouring. Within this mask, CoTracker \citep{karaev2024cotracker} tracks a uniform $20{\times}20$ grid of candidate points across the clip, discarding those outside the mask.
    
    \item \textbf{Motion Trajectories (Fig.~\ref{fig:motion_tracking}.d)} The tracked points produce dense temporal trajectories. A 
    \emph{kinematic filter} 
    selects the top 50\% of points with the largest displacement magnitude, retaining instrument tracks and filtering out background.

    \item \textbf{Motion Embeddings (Fig.~\ref{fig:motion_tracking}.e)} The filtered trajectories (2D coordinates over time, optionally augmented with depth at $t{=}0$) are encoded by a 10-layer LSTM into compact \emph{motion embeddings}. 
\end{itemize}

\vspace{-14pt}
\section{Experimental Setup}
\vspace{-4pt}
\label{sec:experiment_setup}

\begin{table*}[ht!]
\small
\centering

\caption{\textbf{Performance of automated Surgical Action Triplet recognition from surgical video.} Reporting AUCs and relative gains for Instrument, Action, and Tissue prediction across different context inputs with SurgVLP, HecVL, and PeskaVLP base models. We observe consistent gains from added clinical context and further improvements from our \emph{\textbf{temporal instrument tracking}}. Superscripts mark statistically significant AUC gains compared to the previous context condition (paired DeLong test with Holm correction within base model: $^{**}p_{\text{adj}}<0.01$, $^{*}p_{\text{adj}}<0.05$); analysis details in Appendix \ref{apd:stat_tests_table2}.}
\vspace{-7.0pt}
\label{tab:video_iat_auc_gain_results}

\begin{tabular}{llcccccc}
\toprule
\textbf{Base Model} & \textbf{Context} & 
\multicolumn{2}{c}{\textbf{Instrument}} & 
\multicolumn{2}{c}{\textbf{Action}} & 
\multicolumn{2}{c}{\textbf{Tissue}} \\
\cmidrule(lr){3-4} \cmidrule(lr){5-6} \cmidrule(lr){7-8}
& & AUC & Gain & AUC & Gain & AUC & Gain \\

\midrule
GPT-4o & Vision+Proc./Task+Tracking & $0.50_{\pm0.01}$ & - & $0.51_{\pm0.01}$ & - & $0.52_{\pm0.02}$ & - \\

\midrule
\multirow{4}{*}{SurgVLP} 
& Vision & $0.65_{\pm0.02}$ & - & $0.58_{\pm0.02}$ & - & $0.70_{\pm0.01}$ & - \\
& + Procedure & $0.67_{\pm0.02}$ & \uab{3.1\%} & $0.58_{\pm0.02}$ & \dar{0.0\%} & $0.73_{\pm0.01}$ & \uab{4.3\%} \\
& + Task & $0.70_{\pm0.01}$ & \uab{7.7\%} & $0.60^{*}_{\pm0.01}$ & \uab{3.4\%}  & $0.76_{\pm0.02}$ & \uab{8.6\%} \\
& + Temporal Tracking (our) & {\boldmath $0.73_{\pm0.03}$} & \uab{12.3\%} & {\boldmath $0.61_{\pm0.01}$} & \uab{5.2\%} & {\boldmath $0.79^{**}_{\pm0.01}$} & \uab{12.9\%} \\

\midrule
\multirow{4}{*}{HecVL} 
& Vision & $0.68_{\pm0.04}$ & - & $0.60_{\pm0.01}$ & - & $0.74_{\pm0.02}$ & - \\
& + Procedure & $0.68_{\pm0.03}$ & \dar{0.0\%} & $0.60_{\pm0.01}$ & \dar{0.0\%} & $0.76_{\pm0.02}$ & \uab{2.7\%} \\
& + Task & $0.71_{\pm0.04}$ & \uab{4.4\%} & $0.61_{\pm0.01}$ & \uab{1.7\%}  & $0.77_{\pm0.03}$ & \uab{4.1\%} \\
& + Temporal Tracking (our) & {\boldmath $0.74_{\pm0.03}$} & \uab{8.8\%} & {\boldmath $0.62_{\pm0.01}$} & \uab{3.3\%} & {\boldmath $0.77_{\pm0.01}$} & \uab{4.1\%} \\

\midrule    
\multirow{4}{*}{PeskaVLP} 
& Vision & $0.67_{\pm0.03}$ & - & $0.60_{\pm0.02}$ & - & $0.74_{\pm0.02}$ & - \\
& + Procedure & $0.68_{\pm0.03}$ & \uab{1.5\%} & $0.59_{\pm0.02}$ & \dar{1.7\%} & $0.75_{\pm0.02}$ & \uab{1.4\%} \\
& + Task & $0.69_{\pm0.03}$ & \uab{3.0\%} & $0.60_{\pm0.01}$ & \dar{0.0\%} & $0.74_{\pm0.02}$ & \dar{0.0\%} \\
& + Temporal Tracking (our) & {\boldmath $0.74^{*}_{\pm0.03}$} & \uab{10.4\%} & {\boldmath $0.63_{\pm0.02}$} & \uab{5.0\%} & {\boldmath $0.79^{*}_{\pm0.01}$} & \uab{6.8\%} \\
\bottomrule
\end{tabular}

\end{table*}

\begin{table*}[h!]
\small
\centering

\caption{\textbf{Performance of our IAT triplet grounded Surgical Feedback Generation.} Comparison of baselines and our methods. Alignment Score (LLM-as-a-judge) and text generation metrics (WER, ROUGE) are reported. Gains and improvements are relative to the best baseline (GPT-4o, Video+context). Superscripts on Alignment Score indicate statistically significant improvements over the GPT-4o Video+context baseline (stratified Wilcoxon/van Elteren with Holm correction: $^{**}p_{\text{adj}}<.01$, $^{*}p_{\text{adj}}<.05$); see Appendix~\ref{apd:stat_tests_table3}.}
\vspace{-7.0pt}
\label{tab:alignment_text_scores}

\begin{tabular}{llcccccccccc}
\toprule
\textbf{Model} & \textbf{Condition} & 
\multicolumn{4}{c}{\textbf{Alignment Score}} &
\multicolumn{4}{c}{\textbf{Text Generation Scores}} \\
\cmidrule(lr){3-6} \cmidrule(lr){7-10}
& & Mean$\uparrow$ & Gain & $\geq 3$ & $\geq 4$ & 
WER$\downarrow$ & \%imp. & ROUGE$\uparrow$ & \%imp. \\
\midrule
VQA Llama & Video+context & $1.93_{\pm.03}$ &       & 10.1\% & 1.4\% & $11.0_{\pm1.0}$ &        & $0.09_{\pm.02}$ &        \\

SurgVLP$_{+LM}$ & Video+context & $1.84_{\pm.03}$ &  & 3.8\% & 2.9\% & $10.5_{\pm0.7}$ &  & $0.11_{\pm.02}$ & \\

GPT-4o & Video+context & $2.17_{\pm.02}$ &       & 20.6\% & 0.3\% & $5.1_{\pm0.3}$  &        & $0.11_{\pm.01}$ &        \\
\midrule
GPT-4o      & +(I,A,T)   & $2.23^{**}_{\pm.02}$ & \uab{2.8\%} & 25.1\% & 1.7\% & $4.3_{\pm0.2}$ & \dab{15.4\%} & $0.12_{\pm.01}$ & \uab{9.1\%} \\
GPT-4o       & Context+(I,A,T)      & {$2.33^{**}_{\pm.02}$} & \uab{7.4\%} & 32.7\% & 2.4\% & $4.2_{\pm0.2}$ & \dab{18.2\%} & $0.13_{\pm.01}$ & \uab{16.5\%} \\
GPT-4o       & +confidence gate      & {$2.44^{**}_{\pm.03}$} & \uab{12.4\%} & 42.0\% & 2.9\% & $3.5_{\pm0.4}$ & \dab{31.3\%} & $0.18_{\pm.01}$ & \uab{63.6\%} \\
\bottomrule
\end{tabular}
\vspace{-12.0pt}
\end{table*}

All models are trained with a fixed random seed (0) for reproducibility, using GPT-4o (gpt-4o-2024-11-20) for LLM-based tasks (context definitions, triplet extractions, feedback generation) and Gemini 2.5 Pro (gemini-2.5-pro) for ontology construction. Full implementation details, including hyperparameters and specific architectures, are provided in Appendix~\ref{apd:training_details}.

\parahead{Task 1: Video $\rightarrow$ IAT Prediction}
\label{sec:task_1_setup}
We frame IAT prediction as a multi-class classification task with three independent heads (for Instrument, Action, and Tissue) trained on fused multimodal embeddings. The evaluation follows a 5-fold stratified cross-validation scheme to ensure robust performance estimates. We use a two-stage hybrid model for incorporating motion: first, an LSTM is trained as part of a larger fusion architecture to learn a task-relevant motion representation. Then, this trained LSTM component is used as a fixed feature extractor to generate motion embeddings, which are fused with video and text embeddings to train a final MLP classifier.

\noindent\textbf{Conditions:} All variants use identical data splits. Results for both SurgVLP and PeskaVLP backbones are in Table~\ref{tab:video_iat_auc_gain_results}.
\begin{itemize}[leftmargin=*, itemsep=-0.3em, topsep=2pt]
    \item \emph{Vision}: MLP trained on video embeddings only.
    \item \emph{+ Procedure} and \emph{+ Task}: Adds procedure and task text embeddings to the MLP input.
    \item \emph{+ Temporal Tracking (ours)}: Our full, two-stage hybrid model which adds the LSTM-generated motion embeddings.
\end{itemize}

\noindent\textbf{Metrics: }\textit{AUC} per head (Instrument, Action, Tissue), with mean$\pm$STD over 5 folds, which is robust to class imbalance.

\parahead{Task 2: Feedback Generation}
\label{sec:task_2_setup}
For this task, we compare two main approaches: (1) fine-tuning existing vision-language models directly on our video-to-feedback pairs, and (2) few-shot prompting of a state-of-the-art multimodal model (GPT-4o), both with and without our proposed IAT structural conditioning. The fine-tuned baselines explore both a general-domain VLM adapted with LoRA and a surgery-specific video captioning architecture trained end-to-end.

\begin{table*}[ht!]
\centering
\caption{\textbf{Inter-rater agreement on \textit{Alignment Score} for LLM-as-judge.} Based on blind Human ratings of 30 stratified examples of aligned LLM-generated and real-world surgeon provided feedback.}
\label{tab:llm_judge_agreement}
\vspace{-6.0pt}
\begin{tabular}{lcccc}
\toprule
Comparison & $\kappa_{\text{quad}}$ (95\% CI) & $\rho$ (95\% CI) & $P_o$ (95\% CI) \\
\midrule
Human--Human & 0.70 [0.17, 0.92] & 0.58 [0.24, 0.86] & 0.77 [0.60, 0.90] \\
LLM vs.\ Avg.\ Human & 0.82 [0.56, 0.94] & 0.64 [0.33, 0.87] & 0.80 [0.63, 0.93] \\
Human Rater 1 vs.\ LLM & 0.80 [0.54, 0.93] & 0.79 [0.62, 0.93] & 0.77 [0.60, 0.90] \\
Human Rater 2 vs.\ LLM & 0.91 [0.70, 1.00] & 0.88 [0.68, 1.00] & 0.90 [0.80, 1.00] \\
\bottomrule
\end{tabular}
\vspace{-10.0pt}
\end{table*}

\noindent\textbf{Conditions:} Comparisons are shown in Table~\ref{tab:alignment_text_scores}.
\begin{itemize}[leftmargin=*, itemsep=-0.3em, topsep=2pt]
    \item \emph{VQA LLaMA 3.2 (11B)}: A general VLM fine-tuned using Low-Rank Adaptation (LoRA).
    \item \emph{SurgVLP$_{+LM}$}: A surgery-specific video captioning model, fine-tuned end-to-end.
    \item \emph{GPT-4o (video+context)}: A prompted baseline using video and clinical context; no IAT conditioning.
    \item \emph{GPT-4o + IAT}: Our method using IAT-conditioned prompting.
    \item \emph{GPT-4o + IAT + confidence}: Our full method with an uncertainty gate on IAT predictions.
\end{itemize}

\noindent\textbf{Metrics:} \textit{Alignment Score (1–-5)}: A clinician-aligned \emph{LLM-as-judge} score (see Tab.~\ref{tab:llm_judge_rubric}) for semantic alignment. 
Following recent work on LLM-as-judge evaluation, which recommends establishing human inter-rater reliability before using human scores as a reference for LLM evaluators \citep{tam2024framework}, we had two trained raters score a stratified sample and computed quadratic-weighted Cohen's $\kappa$, Spearman's $\rho$, and percent agreement ($P_o$) to validate the rubric and alignment metric.
Uncertainty estimates were using a 2{,}000-sample nonparametric bootstrap.
\textit{WER} and \textit{ROUGE-Recall} are used as secondary scores.

\vspace{-10pt}
\section{Results}
\vspace{-4pt}
\label{sec:results}

\paragraph{Task 1: Video$\rightarrow$IAT Prediction.}
\label{sec:results_task1}
As summarized in Table~\ref{tab:video_iat_auc_gain_results}, adding \emph{procedure/task} context yields modest, consistent gains for \textit{Instrument} and \textit{Tissue} across both backbones, with smaller or mixed changes for \textit{Action}. The \emph{\textbf{largest improvements come from combining context with our temporal tracking}} stream: with \emph{SurgVLP} AUC reaches 0.73/0.61/0.79 for Instrument/Action/Tissue (\,+12.3\%, +5.2\%, +12.9\%), and with \emph{PeskaVLP} 0.74/0.63/0.79 (\,+10.4\%, +5.0\%, +6.8\%). Context helps with recognition of instrument and tissue in feedback delivery, while explicit kinematics provide the bulk of the lift—especially for Instrument and Tissue—leaving Action as the hardest subtask. Confusion matrices are in Fig. \ref{fig:cm_iat_combined} with additional metrics in Appendix \ref{apd:additional_metrics} and additional models tested in Table \ref{tab:video_iat_auc_additional_results}. Statistical analysis is described in Appendix~\ref{apd:stat_tests_table2}.

\vspace{-2pt}
\paragraph{Task 2: Feedback Generation.}
\label{sec:results_task2}

Table~\ref{tab:alignment_text_scores} shows that adding structured conditioning to GPT-4o improves alignment over Video+context baseline. Notably, grounding with the IAT triplet alone (\textit{Fidelity} $2.23\pm0.02$) performs comparably to combining triplets with raw video, while the \textit{Context+(I,A,T)} condition achieves a higher mean \textit{Fidelity} of $2.33\pm0.02$. This suggests that explicit triplet conditioning is more effective than attempting multimodal fusion via GPT-4o with both structured and raw video frame inputs. \emph{\textbf{Our full model with IAT + confidence gating yields the best results.}} Filtering the evaluation set to high-confidence cases (N = 307 down from 1405) further strengthens performance, with a mean \textit{Fidelity} of $2.44\pm0.03$ (+12.4\%). In this subset, 42.0\% of outputs are rated $\geq3$ and 2.9\% rated $\geq4$. Text generation metrics also improve markedly: WER drops by 31.3\% to $3.47\pm0.4$, and ROUGE rises by 63.6\% to $0.18\pm0.01$. All three structured variants yield statistically significant improvements in the ordinal \textit{Fidelity} scores over the GPT-4o Video+context baseline (Appendix~\ref{apd:stat_tests_table3}), with effect sizes increasing from small (IAT-only) to small-to-moderate (Context+IAT+gate). These results highlight that structured, clinically grounded conditioning—when coupled with confidence-based filtering—yields the strongest performance.

\vspace{-6pt}
\subsection{Additional Inspection and Analysis}

\vspace{-2pt}
\paragraph{Qualitative Inspection.}
We manually reviewed generations at the extremes of the fidelity rubric. Perfectly aligned outputs (\textit{score}=5) often arose when the trainer mentioned only a single triplet component (e.g., an action such as ``open''); the model captured this aspect precisely (examples in Table~\ref{tab:iat_feedback_examples}). In contrast, \emph{unsafe} outputs (\textit{score}=1) were rare—\textbf{37/1365} cases—and were overwhelmingly (\textbf{31/37}) triggered by negated trainer instructions (e.g., ``stop'', ``don’t buzz''). Our current IAT representation does not encode negation or ``do-not'' intent, suggesting a useful extension with an explicit \emph{polarity} tag.

\vspace{-2pt}
\paragraph{Calibration and Confidence Gating.}
The Video$\rightarrow$IAT heads exhibit good-to-fair calibration with Expected Calibration Error (ECE) of roughly 0.07, 0.04, and 0.08 percentage points (lower is better). This \emph{\textbf{supports our use of confidence gating}}: conditioning feedback only on high-confidence triplets yields the best results in Table~\ref{tab:alignment_text_scores} (``\,+IAT + confidence\,'').

\vspace{-2pt}
\paragraph{Human Validation of LLM-as-judge.}
We compared LLM and human expert application of the fidelity score. Agreement between LLM and human raters on the ordinal 1--5 fidelity scale was substantial and comparable to human--human agreement in this sample (Table~\ref{tab:llm_judge_agreement}). The Average-Human vs.\ LLM quadratic-weighted Cohen's $\kappa_{\text{quad}} = 0.82$ (95\% CI $[0.56, 0.94]$) indicates reliability well above chance. Residual disagreements were rare and largely off by one point around mid-range scores (2--4), consistent with ordinal judgments.

\vspace{-10pt}
\section{Discussion}
\vspace{-4pt}
\label{sec:discussion}

Our results show that clinically grounded structure enables both recognition and generation: fusing video, procedure/task context, and explicit tool kinematics improves Video$\rightarrow$IAT prediction (Table~\ref{tab:video_iat_auc_gain_results}), and conditioning GPT\mbox{-}4o on the predicted triplets improves \emph{fidelity} of generated feedback beyond strong baselines (Table~\ref{tab:alignment_text_scores}). A score of 3 on our 1–5 rubric denotes \emph{admissible} alignment (captures the trainer’s intent but may miss detail), whereas 5 indicates near-exact match—an idealization that is uncommon even across human trainers \citep{gawad2019inter}.

\vspace{-2pt}
\paragraph{Why a Sparse, Structured Representation?}
We adopt a compact IAT representation because it is \emph{clinically grounded and validated} for modeling tool–tissue interactions in endoscopic/robotic surgery \citep{ nwoye2023cholectriplet2022}; \emph{auditable}, letting each sentence trace to explicit instrument–action–tissue evidence; and \emph{data-efficient} under the heterogeneity of feedback delivery, abstracting away surface variation while preserving core semantics. Crucially, IATs separate \emph{content} (instrument, action, tissue) from \emph{delivery} (tone, hedges, praise): both matter for feedback effectiveness \citep{wong2023development}, but once the core content is correct and inspectable, delivery style can easily be modulated with modern LLMs. Our ontology is induced from real trainer$\to$trainee language, following use of LLMs in medical annotation \citep{goel2023llms},  and normalized with LLM-based clustering (App.~\ref{app:ontology_extraction_details}), aligning labels with clinical usage and mitigating long-tail infrequent variations.

\vspace{-2pt}
\paragraph{Practical Value in Clinical Settings.}
The pipeline supports \emph{post hoc} review: trainees can receive concise, surgeon-style messages linked to concrete tool–tissue events, complementing sparse real-world feedback and helping trainers standardize coaching \citep{ma2024artificial}. The same structure is amenable to \emph{in-room} assistance:  controlled style preserves naturalistic delivery—features known to influence trainee acceptance \citep{kocielnik2024human, knudsen2025mp06}. Such systems may also reduce senior faculty burden during simulation-based practice.

\vspace{-2pt}
\paragraph{Scale and Generalizability of the Data.}
Although our dataset is modest in size and drawn from a single institution, several design choices support generalizability beyond our specific setting. First, we build on pre-trained surgical video/text encoders (e.g., PeskaVLP), adapting representations that were originally trained on broader surgical benchmarks; the strong performance of adapted models suggests that our video data is meaningfully aligned with these wider distributions. Second, we adopt a clinically meaningful sparse triplet representation established in prior work \citep{nwoye2023cholectriplet2022}, reducing the risk of overfitting to an idiosyncratic, task-specific coding scheme. Third, our dataset spans multiple procedures, roughly forty distinct teaching tasks/steps, and a broad set of instrument, action, and tissue mentions, providing coverage across varied surgical contexts rather than a single narrow skill niche.

\vspace{-2pt}
\paragraph{Limitations.}
First, we used LLM-as-a-judge as one of the metrics. While we show high agreement with human judgment on a subset, there is always a risk of systematic drift at scale. Second, because the \emph{polarity} is not explicit in the IAT representation, prohibitions (e.g., ``don't buzz the ureter'') are poorly captured, and most unsafe outputs arise from such negations. Third, IAT representation is \emph{coarse}, degrees of adjustment (``a bit closer''), skill quality, and clinical \emph{criticality} are not represented, which caps the achievable fidelity even with perfect recognition.

\vspace{-2pt}
\paragraph{Future Directions.}
We see three immediate avenues: (1) enrich the IAT representation with tags for \emph{polarity} (prohibit/encourage), \emph{intensity} (degree/magnitude), and \emph{criticality} (clinical importance); (2) model \emph{permissible alternatives} by collecting multiple valid trainer responses for the same context and scoring clinical usefulness rather than alignment; and (3) incorporate constraint-based reasoning to rule out implausible IAT combinations and enforce procedure- and step-specific safety constraints in context.

\vspace{-12pt}
\section{Conclusion}
\vspace{-4pt}
\label{sec:conclusion}
We present the \textit{first} end-to-end system that generates surgeon-aligned, natural-language feedback directly from operative video and clinical context, grounded in explicit Instrument–Action–Tissue (IAT) semantics. Our contributions—(i) a feedback-induced, clinically normalized IAT ontology, (ii) a multimodal video$\rightarrow$IAT predictor that fuses procedure/task context with temporal instrument motion, and (iii) IAT-conditioned GPT-4o with confidence gating and a clinician-aligned evaluation. These yield better alignment while remaining interpretable and auditable, laying the groundwork for scalable surgical coaching, easing workload, and lowering training costs.

\vspace{-6pt}
\acks{Research supported by NCI NIH under Award Numbers R01CA251579 and R01CA298988. The content is solely the responsibility of the authors and does not necessarily represent the official views of the NIH.}

\bibliography{bibliography}

\begin{thebibliography}{45}
\providecommand{\natexlab}[1]{#1}
\providecommand{\url}[1]{\texttt{#1}}
\expandafter\ifx\csname urlstyle\endcsname\relax
  \providecommand{\doi}[1]{doi: #1}\else
  \providecommand{\doi}{doi: \begingroup \urlstyle{rm}\Url}\fi

\bibitem[Agha et~al.(2015)Agha, Fowler, and Sevdalis]{agha2015role}
Riaz~A Agha, Alexander~J Fowler, and Nick Sevdalis.
\newblock The role of non-technical skills in surgery.
\newblock \emph{Annals of medicine and surgery}, 4\penalty0 (4):\penalty0 422--427, 2015.

\bibitem[Assran et~al.(2025)Assran, Bardes, Fan, Garrido, Howes, Muckley, Rizvi, Roberts, Sinha, Zholus, et~al.]{assran2025v}
Mido Assran, Adrien Bardes, David Fan, Quentin Garrido, Russell Howes, Matthew Muckley, Ammar Rizvi, Claire Roberts, Koustuv Sinha, Artem Zholus, et~al.
\newblock V-jepa 2: Self-supervised video models enable understanding, prediction and planning.
\newblock \emph{arXiv preprint arXiv:2506.09985}, 2025.

\bibitem[Babaei~Giglou et~al.(2024)Babaei~Giglou, D’Souza, Engel, and Auer]{babaei2024llms4om}
Hamed Babaei~Giglou, Jennifer D’Souza, Felix Engel, and S{\"o}ren Auer.
\newblock Llms4om: matching ontologies with large language models.
\newblock In \emph{European Semantic Web Conference}, pages 25--35. Springer, 2024.

\bibitem[Bai et~al.(2023)Bai, Islam, Seenivasan, and Ren]{bai2023surgical}
Long Bai, Mobarakol Islam, Lalithkumar Seenivasan, and Hongliang Ren.
\newblock Surgical-vqla: Transformer with gated vision-language embedding for visual question localized-answering in robotic surgery.
\newblock In \emph{2023 IEEE International Conference on Robotics and Automation (ICRA)}, pages 6859--6865. IEEE, 2023.

\bibitem[Bonrath et~al.(2015)Bonrath, Dedy, Gordon, and Grantcharov]{bonrath2015comprehensive}
Esther~M Bonrath, Nicolas~J Dedy, Lauren~E Gordon, and Teodor~P Grantcharov.
\newblock Comprehensive surgical coaching enhances surgical skill in the operating room.
\newblock \emph{Annals of surgery}, 262\penalty0 (2):\penalty0 205--212, 2015.

\bibitem[Chinh et~al.(2019)Chinh, Zade, Ganji, and Aragon]{chinh2019ways}
Bonnie Chinh, Himanshu Zade, Abbas Ganji, and Cecilia Aragon.
\newblock Ways of qualitative coding: A case study of four strategies for resolving disagreements.
\newblock In \emph{Extended abstracts of the 2019 CHI conference on human factors in computing systems}, pages 1--6, 2019.

\bibitem[Comanici et~al.(2025)Comanici, Bieber, Schaekermann, Pasupat, Sachdeva, Dhillon, Blistein, Ram, Zhang, Rosen, et~al.]{comanici2025gemini}
Gheorghe Comanici, Eric Bieber, Mike Schaekermann, Ice Pasupat, Noveen Sachdeva, Inderjit Dhillon, Marcel Blistein, Ori Ram, Dan Zhang, Evan Rosen, et~al.
\newblock Gemini 2.5: Pushing the frontier with advanced reasoning, multimodality, long context, and next generation agentic capabilities.
\newblock \emph{arXiv preprint arXiv:2507.06261}, 2025.

\bibitem[DiMaio et~al.(2011)DiMaio, Hanuschik, and Kreaden]{dimaio2011vinci}
Simon DiMaio, Mike Hanuschik, and Usha Kreaden.
\newblock The da vinci surgical system.
\newblock \emph{Surgical robotics: systems applications and visions}, pages 199--217, 2011.

\bibitem[D’Angelo et~al.(2020)D’Angelo, Ruis, Collier, Shaffer, and Pugh]{d2020evaluating}
Anne-Lise~D D’Angelo, Andrew~R Ruis, Wesley Collier, David~Williamson Shaffer, and Carla~M Pugh.
\newblock Evaluating how residents talk and what it means for surgical performance in the simulation lab.
\newblock \emph{The American Journal of Surgery}, 220\penalty0 (1):\penalty0 37--43, 2020.

\bibitem[Gawad et~al.(2019)Gawad, Fowler, Mimeault, and Raiche]{gawad2019inter}
Nada Gawad, Amanda Fowler, Richard Mimeault, and Isabelle Raiche.
\newblock The inter-rater reliability of technical skills assessment and retention of rater training.
\newblock \emph{Journal of surgical education}, 76\penalty0 (4):\penalty0 1088--1093, 2019.

\bibitem[Goel et~al.(2023)Goel, Gueta, Gilon, Liu, Erell, Nguyen, Hao, Jaber, Reddy, Kartha, et~al.]{goel2023llms}
Akshay Goel, Almog Gueta, Omry Gilon, Chang Liu, Sofia Erell, Lan~Huong Nguyen, Xiaohong Hao, Bolous Jaber, Shashir Reddy, Rupesh Kartha, et~al.
\newblock Llms accelerate annotation for medical information extraction.
\newblock In \emph{machine learning for health (ML4H)}, pages 82--100. PMLR, 2023.

\bibitem[Gupta et~al.(2025)Gupta, Kocielnik, Wang, Nasriddinov, Yang, Wong, Anandkumar, and Hung]{gupta2025multi}
Arushi Gupta, Rafal~Dariusz Kocielnik, Jiayun Wang, Firdavs Nasriddinov, Cherine Yang, Elyssa Wong, Anima Anandkumar, and Andrew Hung.
\newblock Multi-modal self-supervised learning for surgical feedback effectiveness assessment.
\newblock In \emph{Machine Learning for Health (ML4H)}, pages 440--455. PMLR, 2025.

\bibitem[Haque et~al.(2022)Haque, Hui, You, Ma, Nguyen, Lei, Cen, Aron, Collins, Djaladat, et~al.]{haque2022assessment}
Taseen~F Haque, Alvin Hui, Jonathan You, Runzhuo Ma, Jessica~H Nguyen, Xiaomeng Lei, Steven Cen, Monish Aron, Justin~W Collins, Hooman Djaladat, et~al.
\newblock An assessment tool to provide targeted feedback to robotic surgical trainees: development and validation of the end-to-end assessment of suturing expertise (ease).
\newblock \emph{Urology practice}, 9\penalty0 (6):\penalty0 532--539, 2022.

\bibitem[Karaev et~al.(2024)Karaev, Rocco, Graham, Neverova, Vedaldi, and Rupprecht]{karaev2024cotracker}
Nikita Karaev, Ignacio Rocco, Benjamin Graham, Natalia Neverova, Andrea Vedaldi, and Christian Rupprecht.
\newblock Cotracker: It is better to track together.
\newblock In \emph{European conference on computer vision}, pages 18--35. Springer, 2024.

\bibitem[Kiyasseh et~al.(2023)Kiyasseh, Ma, Haque, Miles, Wagner, Donoho, Anandkumar, and Hung]{kiyasseh2023vision}
Dani Kiyasseh, Runzhuo Ma, Taseen~F Haque, Brian~J Miles, Christian Wagner, Daniel~A Donoho, Animashree Anandkumar, and Andrew~J Hung.
\newblock A vision transformer for decoding surgeon activity from surgical videos.
\newblock \emph{Nature biomedical engineering}, 7\penalty0 (6):\penalty0 780--796, 2023.

\bibitem[Knudsen et~al.(2025)Knudsen, Kocielnik, Wong, Lin, Shiang, Lin, Goldenberg, Lee, and Hung]{knudsen2025mp06}
J~Everett Knudsen, Rafal Kocielnik, Elyssa~Y Wong, Jasmine Lin, Alex Shiang, Lydia Lin, Mitchell Goldenberg, Randall~A Lee, and Andrew~J Hung.
\newblock Mp06-15 ai discovery of surgical feedback quality: A first step towards automated real-time delivery of feedback.
\newblock \emph{Journal of Urology}, 213\penalty0 (5S):\penalty0 e148, 2025.

\bibitem[Kocielnik et~al.(2023)Kocielnik, Wong, Chu, Lin, Huang, Wang, Anandkumar, and Hung]{kocielnik2023deep}
Rafal Kocielnik, Elyssa~Y Wong, Timothy~N Chu, Lydia Lin, De-An Huang, Jiayun Wang, Anima Anandkumar, and Andrew~J Hung.
\newblock Deep multimodal fusion for surgical feedback classification.
\newblock In \emph{Machine Learning for Health (ML4H)}, pages 256--267. PMLR, 2023.

\bibitem[Kocielnik et~al.(2024)Kocielnik, Yang, Ma, Cen, Wong, Chu, Knudsen, Wager, Heard, Ghaffar, et~al.]{kocielnik2024human}
Rafal Kocielnik, Cherine~H Yang, Runzhuo Ma, Steven~Y Cen, Elyssa~Y Wong, Timothy~N Chu, J~Everett Knudsen, Peter Wager, John Heard, Umar Ghaffar, et~al.
\newblock Human ai collaboration for unsupervised categorization of live surgical feedback.
\newblock \emph{NPJ Digital Medicine}, 7\penalty0 (1):\penalty0 372, 2024.

\bibitem[Laca et~al.(2022)Laca, Kocielnik, Nguyen, You, Tsang, Wong, Shtulman, Anandkumar, and Hung]{laca2022using}
Jasper~A Laca, Rafal Kocielnik, Jessica~H Nguyen, Jonathan You, Ryan Tsang, Elyssa~Y Wong, Andrew Shtulman, Anima Anandkumar, and Andrew~J Hung.
\newblock Using real-time feedback to improve surgical performance on a robotic tissue dissection task.
\newblock \emph{European Urology Open Science}, 46:\penalty0 15--21, 2022.

\bibitem[Li et~al.(2023)Li, Xia, Luo, He, and Jia]{li2023mt}
Yuchong Li, Tong Xia, Huoling Luo, Baochun He, and Fucang Jia.
\newblock Mt-fist: a multi-task fine-grained spatial-temporal framework for surgical action triplet recognition.
\newblock \emph{IEEE journal of biomedical and health informatics}, 27\penalty0 (10):\penalty0 4983--4994, 2023.

\bibitem[Low et~al.(2025)Low, Wang, Zhang, Zeng, Zhuo, Mazomenos, and Jin]{low2025surgraw}
Chang~Han Low, Ziyue Wang, Tianyi Zhang, Zhitao Zeng, Zhu Zhuo, Evangelos~B Mazomenos, and Yueming Jin.
\newblock Surgraw: Multi-agent workflow with chain-of-thought reasoning for surgical intelligence.
\newblock \emph{arXiv preprint arXiv:2503.10265}, 2025.

\bibitem[Ma et~al.(2022)Ma, Ramaswamy, Xu, Trinh, Kiyasseh, Chu, Wong, Lee, Rodriguez, DeMeo, et~al.]{ma2022surgical}
Runzhuo Ma, Ashwin Ramaswamy, Jiashu Xu, Loc Trinh, Dani Kiyasseh, Timothy~N Chu, Elyssa~Y Wong, Ryan~S Lee, Ivan Rodriguez, Gina DeMeo, et~al.
\newblock Surgical gestures as a method to quantify surgical performance and predict patient outcomes.
\newblock \emph{NPJ Digital Medicine}, 5\penalty0 (1):\penalty0 187, 2022.

\bibitem[Ma et~al.(2024)Ma, Kiyasseh, Laca, Kocielnik, Wong, Chu, Cen, Yang, Dalieh, Haque, et~al.]{ma2024artificial}
Runzhuo Ma, Dani Kiyasseh, Jasper~A Laca, Rafal Kocielnik, Elyssa~Y Wong, Timothy~N Chu, Steven Cen, Cherine~H Yang, Istabraq~S Dalieh, Taseen~F Haque, et~al.
\newblock Artificial intelligence-based video feedback to improve novice performance on robotic suturing skills: a pilot study.
\newblock \emph{Journal of Endourology}, 38\penalty0 (8):\penalty0 884--891, 2024.

\bibitem[Nwoye et~al.(2020)Nwoye, Gonzalez, Yu, Mascagni, Mutter, Marescaux, and Padoy]{nwoye2020recognition}
Chinedu~Innocent Nwoye, Cristians Gonzalez, Tong Yu, Pietro Mascagni, Didier Mutter, Jacques Marescaux, and Nicolas Padoy.
\newblock Recognition of instrument-tissue interactions in endoscopic videos via action triplets.
\newblock In \emph{International conference on medical image computing and computer-assisted intervention}, pages 364--374. Springer, 2020.

\bibitem[Nwoye et~al.(2022)Nwoye, Yu, Gonzalez, Seeliger, Mascagni, Mutter, Marescaux, and Padoy]{nwoye2022rendezvous}
Chinedu~Innocent Nwoye, Tong Yu, Cristians Gonzalez, Barbara Seeliger, Pietro Mascagni, Didier Mutter, Jacques Marescaux, and Nicolas Padoy.
\newblock Rendezvous: Attention mechanisms for the recognition of surgical action triplets in endoscopic videos.
\newblock \emph{Medical Image Analysis}, 78:\penalty0 102433, 2022.

\bibitem[Nwoye et~al.(2023{\natexlab{a}})Nwoye, Alapatt, Yu, Vardazaryan, Xia, Zhao, Xia, Jia, Yang, Wang, et~al.]{nwoye2023cholectriplet2021}
Chinedu~Innocent Nwoye, Deepak Alapatt, Tong Yu, Armine Vardazaryan, Fangfang Xia, Zixuan Zhao, Tong Xia, Fucang Jia, Yuxuan Yang, Hao Wang, et~al.
\newblock Cholectriplet2021: A benchmark challenge for surgical action triplet recognition.
\newblock \emph{Medical Image Analysis}, 86:\penalty0 102803, 2023{\natexlab{a}}.

\bibitem[Nwoye et~al.(2023{\natexlab{b}})Nwoye, Yu, Sharma, Murali, Alapatt, Vardazaryan, Yuan, Hajek, Reiter, Yamlahi, et~al.]{nwoye2023cholectriplet2022}
Chinedu~Innocent Nwoye, Tong Yu, Saurav Sharma, Aditya Murali, Deepak Alapatt, Armine Vardazaryan, Kun Yuan, Jonas Hajek, Wolfgang Reiter, Amine Yamlahi, et~al.
\newblock Cholectriplet2022: Show me a tool and tell me the triplet—an endoscopic vision challenge for surgical action triplet detection.
\newblock \emph{Medical Image Analysis}, 89:\penalty0 102888, 2023{\natexlab{b}}.

\bibitem[Philipp et~al.(2022)Philipp, Alperovich, Gutt-Will, Mathis, Saur, Raabe, and Mathis-Ullrich]{philipp2022dynamic}
Markus Philipp, Anna Alperovich, Marielena Gutt-Will, Andrea Mathis, Stefan Saur, Andreas Raabe, and Franziska Mathis-Ullrich.
\newblock Dynamic cnns using uncertainty to overcome domain generalization for surgical instrument localization.
\newblock In \emph{Proceedings of the IEEE/CVF Winter Conference on Applications of Computer Vision}, pages 3612--3621, 2022.

\bibitem[Romano et~al.(2006)Romano, Kromrey, Coraggio, and Skowronek]{romano2006appropriate}
Jeanine Romano, Jeffrey~D Kromrey, Jesse Coraggio, and Jeff Skowronek.
\newblock Appropriate statistics for ordinal level data: Should we really be using t-test and cohen’sd for evaluating group differences on the nsse and other surveys.
\newblock In \emph{annual meeting of the Florida Association of Institutional Research}, volume 177, 2006.

\bibitem[Schmidgall et~al.(2024{\natexlab{a}})Schmidgall, Cho, Zakka, and Hiesinger]{schmidgall2024gp}
Samuel Schmidgall, Joseph Cho, Cyril Zakka, and William Hiesinger.
\newblock Gp-vls: A general-purpose vision language model for surgery.
\newblock \emph{arXiv preprint arXiv:2407.19305}, 2024{\natexlab{a}}.

\bibitem[Schmidgall et~al.(2024{\natexlab{b}})Schmidgall, Kim, Jopling, and Krieger]{schmidgall2024general}
Samuel Schmidgall, Ji~Woong Kim, Jeffery Jopling, and Axel Krieger.
\newblock General surgery vision transformer: A video pre-trained foundation model for general surgery.
\newblock \emph{arXiv preprint arXiv:2403.05949}, 2024{\natexlab{b}}.

\bibitem[Seenivasan et~al.(2022)Seenivasan, Islam, Krishna, and Ren]{seenivasan2022surgical}
Lalithkumar Seenivasan, Mobarakol Islam, Adithya~K Krishna, and Hongliang Ren.
\newblock Surgical-vqa: Visual question answering in surgical scenes using transformer.
\newblock In \emph{International Conference on Medical Image Computing and Computer-Assisted Intervention}, pages 33--43. Springer, 2022.

\bibitem[Seenivasan et~al.(2023)Seenivasan, Islam, Kannan, and Ren]{seenivasan2023surgicalgpt}
Lalithkumar Seenivasan, Mobarakol Islam, Gokul Kannan, and Hongliang Ren.
\newblock Surgicalgpt: end-to-end language-vision gpt for visual question answering in surgery.
\newblock In \emph{International conference on medical image computing and computer-assisted intervention}, pages 281--290. Springer, 2023.

\bibitem[Servais et~al.(2025)Servais, Rashidi, Porwal, Garibaldi, and Hung]{servais2025novel}
Elliot~L Servais, Laila Rashidi, Priyanshi Porwal, Mark Garibaldi, and Andrew~J Hung.
\newblock Novel force feedback technology improves suturing in robotic-assisted surgery: a pre-clinical study.
\newblock \emph{Surgical Endoscopy}, 39\penalty0 (2):\penalty0 1217--1226, 2025.

\bibitem[Sharma et~al.(2023)Sharma, Nwoye, Mutter, and Padoy]{sharma2023surgical}
Saurav Sharma, Chinedu~Innocent Nwoye, Didier Mutter, and Nicolas Padoy.
\newblock Surgical action triplet detection by mixed supervised learning of instrument-tissue interactions.
\newblock In \emph{International Conference on Medical Image Computing and Computer-Assisted Intervention}, pages 505--514. Springer, 2023.

\bibitem[Tam et~al.(2024)Tam, Sivarajkumar, Kapoor, Stolyar, Polanska, McCarthy, Osterhoudt, Wu, Visweswaran, Fu, et~al.]{tam2024framework}
Thomas Yu~Chow Tam, Sonish Sivarajkumar, Sumit Kapoor, Alisa~V Stolyar, Katelyn Polanska, Karleigh~R McCarthy, Hunter Osterhoudt, Xizhi Wu, Shyam Visweswaran, Sunyang Fu, et~al.
\newblock A framework for human evaluation of large language models in healthcare derived from literature review.
\newblock \emph{NPJ digital medicine}, 7\penalty0 (1):\penalty0 258, 2024.

\bibitem[Terragni et~al.(2021)Terragni, Fersini, Galuzzi, Tropeano, and Candelieri]{terragni2020octis}
Silvia Terragni, Elisabetta Fersini, Bruno~Giovanni Galuzzi, Pietro Tropeano, and Antonio Candelieri.
\newblock {OCTIS}: Comparing and optimizing topic models is simple!
\newblock In \emph{Proceedings of the 16th Conference of the European Chapter of the Association for Computational Linguistics: System Demonstrations}, pages 263--270. Association for Computational Linguistics, April 2021.
\newblock URL \url{https://www.aclweb.org/anthology/2021.eacl-demos.31}.

\bibitem[Tong et~al.(2022)Tong, Song, Wang, and Wang]{tong2022videomae}
Zhan Tong, Yibing Song, Jue Wang, and Limin Wang.
\newblock Videomae: Masked autoencoders are data-efficient learners for self-supervised video pre-training.
\newblock \emph{Advances in neural information processing systems}, 35:\penalty0 10078--10093, 2022.

\bibitem[Wang et~al.(2024)Wang, Bai, Nah, Wang, Zhang, Chen, Wu, Islam, Liu, and Ren]{wang2024surgical}
Guankun Wang, Long Bai, Wan~Jun Nah, Jie Wang, Zhaoxi Zhang, Zhen Chen, Jinlin Wu, Mobarakol Islam, Hongbin Liu, and Hongliang Ren.
\newblock Surgical-lvlm: Learning to adapt large vision-language model for grounded visual question answering in robotic surgery.
\newblock \emph{arXiv preprint arXiv:2405.10948}, 2024.

\bibitem[Wong et~al.(2023)Wong, Chu, Ma, Dalieh, Yang, Ramaswamy, Medina, Kocielnik, Ladi-Seyedian, Shtulman, et~al.]{wong2023development}
Elyssa~Y Wong, Timothy~N Chu, Runzhuo Ma, Istabraq~S Dalieh, Cherine~H Yang, Ashwin Ramaswamy, Luis~G Medina, Rafal Kocielnik, Seyedeh-Sanam Ladi-Seyedian, Andrew Shtulman, et~al.
\newblock Development of a classification system for live surgical feedback.
\newblock \emph{JAMA Network Open}, 6\penalty0 (6):\penalty0 e2320702--e2320702, 2023.

\bibitem[Yang et~al.(2024{\natexlab{a}})Yang, Kang, Huang, Xu, Feng, and Zhao]{yang2024depth}
Lihe Yang, Bingyi Kang, Zilong Huang, Xiaogang Xu, Jiashi Feng, and Hengshuang Zhao.
\newblock Depth anything: Unleashing the power of large-scale unlabeled data.
\newblock In \emph{Proceedings of the IEEE/CVF conference on computer vision and pattern recognition}, pages 10371--10381, 2024{\natexlab{a}}.

\bibitem[Yang et~al.(2024{\natexlab{b}})Yang, Zhao, Xu, Qi, Lu, Phung, and Du]{yang2024neural}
Xiaohao Yang, He~Zhao, Weijie Xu, Yuanyuan Qi, Jueqing Lu, Dinh Phung, and Lan Du.
\newblock Neural topic modeling with large language models in the loop.
\newblock \emph{arXiv preprint arXiv:2411.08534}, 2024{\natexlab{b}}.

\bibitem[Yuan et~al.(2023)Yuan, Srivastav, Yu, Lavanchy, Mascagni, Navab, and Padoy]{yuan2023learning}
Kun Yuan, Vinkle Srivastav, Tong Yu, Joel Lavanchy, Pietro Mascagni, Nassir Navab, and Nicolas Padoy.
\newblock Learning multi-modal representations by watching hundreds of surgical video lectures.
\newblock \emph{arXiv preprint arXiv:2307.15220}, 2023.

\bibitem[Yuan et~al.(2024)Yuan, Navab, Padoy, et~al.]{yuan2024procedure}
Kun Yuan, Nassir Navab, Nicolas Padoy, et~al.
\newblock Procedure-aware surgical video-language pretraining with hierarchical knowledge augmentation.
\newblock \emph{Advances in Neural Information Processing Systems}, 37:\penalty0 122952--122983, 2024.

\bibitem[Yuan et~al.(2025)Yuan, Srivastav, Yu, Lavanchy, Marescaux, Mascagni, Navab, and Padoy]{yuan2025learning}
Kun Yuan, Vinkle Srivastav, Tong Yu, Joel~L Lavanchy, Jacques Marescaux, Pietro Mascagni, Nassir Navab, and Nicolas Padoy.
\newblock Learning multi-modal representations by watching hundreds of surgical video lectures.
\newblock \emph{Medical Image Analysis}, page 103644, 2025.

\end{thebibliography}

\appendix

\section{Fidelity Scoring Rubric (LLM-as-a-Judge)}
\label{app:fidelity_rubric}

This appendix specifies the 1--5 \emph{Fidelity} scale used to judge how well a generated message aligns with a trainer’s reference. The evaluator (GPT\mbox{-}4o) compares \emph{only} the requested tool--tissue action(s), ignoring phrasing/style, and outputs a single integer score. The levels below capture semantic/clinical agreement from unsafe/opposite (1) to exact match in intent, action, and target (5). See Table~\ref{tab:llm_judge_rubric}.

\begin{table*}[h!]
\small
\centering
\setlength{\tabcolsep}{6pt}
\renewcommand{\arraystretch}{1.25}
\caption{\textbf{Alignment Score: LLM-as-a-judge scoring rubric for judging alignment between generated vs.\ trainer provided feedback.} 
Evaluator compares the generated message to the expert reference and scores \emph{only} the requested action(s) on a 1--5 fidelity scale.}
\label{tab:llm_judge_rubric}
\begin{tabular}{
  >{\raggedright\arraybackslash}p{0.27\textwidth}
  >{\raggedright\arraybackslash}p{0.32\textwidth}
  >{\raggedright\arraybackslash}p{0.34\textwidth}
}
\toprule
\textbf{Score} & \textbf{Definition} & \textbf{Examples} \\
\midrule
\textbf{1 — Opposite or Unsafe Action} &
Feedback suggests an action that is opposite in intent or clearly unsafe compared with the ground truth. &
``cut the vein'' vs.\ ``secure the vein''; ``stop'' vs.\ ``continue''; ``only one clip'' vs.\ ``put all the clips'' \\
\addlinespace[2pt]
\textbf{2 — Wrong or Mismatched Action} &
Feedback conveys a different type or modality of action. &
``sweep towards you'' vs.\ ``pull the tissue''; ``buzz the artery'' vs.\ ``clip the artery'' \\
\addlinespace[2pt]
\textbf{3 — Partially Aligned, Missing/Extra Detail} &
Matches the general action but adds or omits key details (amount, precision, target tissue, force). &
``stop the bleed'' vs.\ ``buzz that bleeder''; ``clip to the artery'' vs.\ ``only one clip to artery'' \\
\addlinespace[2pt]
\textbf{4 — Mostly Aligned, Minor Wording Differences} &
Core action and target match; differences are hedging, emphasis, or style only. &
``cauterize this'' vs.\ ``buzz that bleeder''; ``closer to prostate'' vs.\ ``come 1\,mm closer to the prostate'' \\
\addlinespace[2pt]
\textbf{5 — Perfectly Aligned} &
Exactly matches intent, action, and target tissue/instrument. &
``coag the vein'' vs.\ ``buzz that bleeder''; ``move the left hand under the ureter'' vs.\ ``get L hand below ureter'' \\
\bottomrule
\end{tabular}
\end{table*}

\section{Ontology Induction and Clustering Benchmarks}
\label{app:ontology_extraction_details}

To normalize free-form trainer $\rightarrow$ trainee language into a clinically usable label space, we cluster surface forms for \emph{Instrument}, \emph{Action}, and \emph{Tissue} mentions and map them to canonical tags. We compared three clustering pipelines using an embedding-based cluster coherence metric implementation from \cite{terragni2020octis} (higher is better): (i) an unsupervised BERTopic baseline, (ii) an initial LLM-driven clustering, which used a zero-shot prompt to group raw textual mentions into semantically related clusters without domain-specific examples, and (iii) a refined LLM-driven variant (\emph{Final LLM clustering}). The refined variant takes the initial LLM mappings and few-shot examples as input and merges overly nuanced clusters into more general, functionally relevant categories. This two-step approach ensures that the final label space is robust to minor lexical variations and contains categories frequent enough for effective model training, while still preserving core clinical meaning. As shown in Table~\ref{tab:clustering_coherence}, the final approach yields the highest coherence for \textit{Action} and \textit{Tissue} (+0.18 and +0.36 over baseline, respectively) and competitive \textit{Instrument} coherence, so we adopt it to build the feedback-induced IAT dictionary used throughout our modeling pipeline (Sec.~\ref{sec:pipeline}).

\begin{table*}[h]
\centering
\small
\begin{tabular}{lcccccc}
\toprule
\textbf{Method} & \textbf{Instr.} & $\Delta$ & \textbf{Action} & $\Delta$ & \textbf{Tissue} & $\Delta$ \\
\midrule
BERTopic clustering & 0.49 & 0.00 & 0.56 & 0.00 & 0.50 & 0.00 \\
Initial LLM clustering & \textbf{0.58} & +0.18 & 0.59 & +0.05 & 0.55 & +0.10 \\
Final LLM clustering & 0.55 & +0.12 & \textbf{0.66} & +0.18 & \textbf{0.68} & +0.36 \\
\bottomrule
\end{tabular}
\caption{\textbf{Evaluation of clustering methods for grouping the individual IAT mentions mined from feedback text.} Cluster coherence (higher is better) for ontology induction across the three components of the surgical action triplet. $\Delta$ denotes absolute improvement over the BERTopic baseline. We adopt the  \emph{Final LLM clustering} in the main pipeline.}
\label{tab:clustering_coherence}
\end{table*}

\section{Concrete LLM Prompts}
\label{app:llmprompts}

\subsection{GPT-4o Prompt for Surgical Feedback Generation}
\label{app:gpt4o_fbk_gen_prompt}
We provide the prompt used to generate surgeon-aligned natural-language feedback conditioned on \emph{structured knowledge} extracted from the surgical scene. Concretely, the prompt (Table~\ref{tab:feedback_prompt}) conditions GPT-4o on predicted \emph{Instrument–Action–Tissue (IAT)} triplets, procedure and task summaries, optional class definitions, a few reference (IAT$\rightarrow$feedback) examples, and short frame sequences. The model is instructed to return a single, concise (1–3 sentence) trainee-directed message without boilerplate. At inference, only triplets that pass our calibrated confidence gate are injected; otherwise a no-evidence placeholder is used. We apply this same prompt across all experiments reported in the paper.

\begin{table*}[t]
  \captionsetup{aboveskip=4pt,belowskip=2pt} 
  \small
  \centering
  \caption{\textbf{Prompt for surgeon-aligned feedback generation from video and IAT context.}}
  \label{tab:feedback_prompt}
  \begin{tcolorbox}[
    enhanced, 
    width=\textwidth,
    colback=white,
    colframe=black!75!white,
    colbacktitle=black!85!white,
    coltitle=white,
    title=Feedback Generation Prompt,
    fonttitle=\bfseries,
    boxrule=0.4pt,
    arc=0pt, outer arc=0pt,    left=3pt,right=3pt,top=3pt,bottom=3pt,boxsep=2pt,
    before skip=2pt, after skip=2pt
  ]
  \footnotesize 

  \textbf{Role.} You are an expert surgical training assistant specializing in urological procedures performed with the da Vinci surgical system. Your primary function is to analyze surgical scene data—video frames and abstract triplets—and generate concise, actionable, pedagogical feedback for a surgeon in training.

  \medskip
  \textbf{Detailed instructions}
  \begin{enumerate}[leftmargin=*,nosep,label=\arabic*.]
    \item Analyze the visual context: assess instrument position, motion, and tissue state.
    \item Interpret the core event: identify the key tool–tissue interaction.
    \item Consult the lexicon: respect class definitions and terminology.
    \item Situate the task: tailor guidance to the current task and procedure.
    \item Learn from precedent: use (IAT$\rightarrow$Feedback) examples when relevant.
    \item Synthesize actionable feedback: state what to do and why.
    \item Ensure feedback is constructive: specific, supportive, and safety-oriented.
    \item Maintain focus: address the observed event; avoid digressions.
    \item Reduce redundancy: avoid repeating obvious details.
  \end{enumerate}
    \medskip
  \textbf{Formatting instructions}
  \begin{itemize}[leftmargin=*,nosep]
    \item Output a single concise string (1–3 sentences).
    \item Do not include prefixes (e.g., “Feedback:”), bullets, or extra explanations.
    \item Tone: professional, direct, educational.
  \end{itemize}
\medskip
  \textbf{Inputs}
  \begin{itemize}[leftmargin=*,nosep]
    \item \textit{Video Frames} (list of images): sequential frames from the surgical clip.
    \item \textit{IAT Triplet} (tuple): (Instrument, Action, Tissue) for the event.
    \item \textit{Class Definitions} (dict, optional): descriptions for instrument/action/tissue classes.
    \item \textit{Procedure} (str) and \textit{Procedure Definition} (str).
    \item \textit{Task} (str) and \textit{Task Definition} (str).
    \item \textit{Reference Examples} (list): (IAT$\rightarrow$Feedback) pairs.
  \end{itemize}
\medskip
  \textbf{Output}
    \begin{itemize}[leftmargin=*,nosep]
    \item (str) A single string containing actionable surgical feedback.
  \end{itemize}
  \medskip
  {\ttfamily
  Procedure: \{procedure\}\\
  Procedure Definition: \{procedure\_defn\}\\
  Task: \{task\}\\
  Task Definition: \{task\_defn\}\\
  Observed Event (IAT Triplet): \{iat\_triplet\}\\
  Class Definitions: \{class\_definitions\}\\
  Reference Examples: \{reference\_examples\}\\
  Video Frames: \{video\_frames\}\\
  Actionable Feedback:
  }
  \end{tcolorbox}
\end{table*}

\begin{table*}[t]
  \captionsetup{aboveskip=4pt,belowskip=2pt} 
  \small
  \centering
  \caption{\textbf{Prompt for generating concise definitions of urologic procedures.}}
  \label{tab:procedure_definition_prompt}
  \begin{tcolorbox}[
    enhanced, 
    width=\textwidth,
    colback=white,
    colframe=black!75!white,
    colbacktitle=black!85!white,
    coltitle=white,
    title=Procedure Definition Prompt,
    fonttitle=\bfseries,
    boxrule=0.4pt,
    arc=0pt, outer arc=0pt,
    left=3pt,right=3pt,top=3pt,bottom=3pt,boxsep=2pt,
    before skip=2pt, after skip=2pt
  ]
  \footnotesize

  \textbf{Role.} You are a urologic surgery educator writing a clear, neutral definition of the named procedure.

  \medskip
  \textbf{Instructions}
  \begin{enumerate}[leftmargin=*,nosep,label=\arabic*.]
    \item State the purpose/indications in one sentence.
    \item Describe the essential operative scope (what is removed, repaired, or reconstructed).
    \item Note common approach variants (e.g., open, laparoscopic, robotic, endoscopic) without steps.
    \item Optionally mention a high-level recovery or outcome consideration in one short clause.
    \item Keep it factual, jargon-light, and generalizable.
  \end{enumerate}
    \medskip
  \textbf{Formatting}
  \begin{itemize}[leftmargin=*,nosep]
    \item Output exactly one paragraph, 3--5 sentences, no lists or prefixes.
    \item Tone: professional, neutral, educational.
  \end{itemize}
    \medskip
  \textbf{Input}
  \begin{itemize}[leftmargin=*,nosep]
    \item \textit{Procedure Name} (str; urology).
  \end{itemize}
    \medskip
  \textbf{Output}
  \begin{itemize}[leftmargin=*,nosep]
    \item (str) Single-paragraph definition covering purpose, scope, approach variants, and (optionally) recovery/outcome.
  \end{itemize}
    \medskip
  {\ttfamily
  Procedure Name: \{procedure\_name\}\\[2pt]
  Definition:
  }
  \end{tcolorbox}
\end{table*}

\begin{table*}[t]
  \captionsetup{aboveskip=4pt,belowskip=2pt} 
  \small
  \centering
  \caption{\textbf{Prompt for generating concise definitions of urologic surgical tasks.}}
  \label{tab:task_definition_prompt}
  \begin{tcolorbox}[
    enhanced, 
    width=\textwidth,
    colback=white,
    colframe=black!75!white,
    colbacktitle=black!85!white,
    coltitle=white,
    title=Task Definition Prompt,
    fonttitle=\bfseries,
    boxrule=0.4pt,
    arc=0pt, outer arc=0pt,
    left=3pt,right=3pt,top=3pt,bottom=3pt,boxsep=2pt,
    before skip=2pt, after skip=2pt
  ]
  \footnotesize

  \textbf{Role.} You are a urologic surgery educator writing a clear, neutral definition of the named teaching \emph{task}.

  \medskip
  \textbf{Instructions}
  \begin{enumerate}[leftmargin=*,nosep,label=\arabic*.]
    \item State the task’s purpose in one sentence.
    \item Ground it in key anatomy or structures involved.
    \item Describe the core competency (what the trainee must reliably achieve).
    \item Mention a salient safety or preservation consideration.
    \item Expand abbreviations on first use when ambiguous.
  \end{enumerate}
    \medskip
  \textbf{Formatting}
  \begin{itemize}[leftmargin=*,nosep]
    \item Output exactly one paragraph, 2--4 sentences, no lists or prefixes.
    \item Tone: professional, neutral, educational.
  \end{itemize}
    \medskip
  \textbf{Input}
  \begin{itemize}[leftmargin=*,nosep]
    \item \textit{Task Name} (str; urology).
  \end{itemize}
    \medskip
  \textbf{Output}
  \begin{itemize}[leftmargin=*,nosep]
    \item (str) Single-paragraph task definition covering purpose, anatomy, core competency, and a key safety point.
  \end{itemize}
    \medskip
  {\ttfamily
  Task Name: \{task\_name\}\\[2pt]
  Task Definition:
  }
  \end{tcolorbox}
\end{table*}

\begin{table*}[t]
  \captionsetup{aboveskip=4pt,belowskip=2pt} 
  \small
  \centering
  \caption{\textbf{Prompt for extracting \emph{instrument} component of triplet from a feedback line.}}
  \label{tab:instrument_extraction_prompt}
  \begin{tcolorbox}[
    enhanced, 
    width=\textwidth,
    colback=white,
    colframe=black!75!white,
    colbacktitle=black!85!white,
    coltitle=white,
    title=Instrument Extraction Prompt,
    fonttitle=\bfseries,
    boxrule=0.4pt,
    arc=0pt, outer arc=0pt,
    left=3pt,right=3pt,top=3pt,bottom=3pt,boxsep=2pt,
    before skip=2pt, after skip=2pt
  ]
  \footnotesize

  \textbf{Instruction.} You are working in the context of verbal feedback delivered by a trainer to a trainee in a live robot-assystec daVinci surgery. Extract the following aspects from given feedback line:\\
  instrument or tool - identify any mentions of surgical instrument or tool. If no instrument or tool is mentioned, produce \"NONE\" \\
  Make sure you extract all and every surgical instruments or tools including robotic daVinci instruments, make sure to extract mentions of instruments or tools such as catheter, needle, stitches, sutures, robotic arms, graspers, forcepts, retractor, spreader, electrocautery, energy devices, camera, left hand, right hand, 4th arm or others. \\
  Also extract instruments and tools that are strongly implied, even if not mentioned explicitly, based on the action context. \\
  Produce output strictly in the format [instrument 1, instrument 2, ...].

  \medskip
  
  {\ttfamily
  Process this feedback line: \{feedback\_line\}
  }
  \end{tcolorbox}
\end{table*}

\begin{table*}[t]
  \captionsetup{aboveskip=4pt,belowskip=2pt} 
  \small
  \centering
  \caption{\textbf{Prompt for extracting \emph{action} component of triplet from a feedback line.}}
  \label{tab:action_extraction_prompt}
  \begin{tcolorbox}[
    enhanced, 
    width=\textwidth,
    colback=white,
    colframe=black!75!white,
    colbacktitle=black!85!white,
    coltitle=white,
    title=Action Extraction Prompt,
    fonttitle=\bfseries,
    left=3pt,right=3pt,top=3pt,bottom=3pt,boxsep=2pt,
    before skip=2pt, after skip=2pt
  ]
  \footnotesize

  \textbf{Instruction.} You are working in the context of verbal feedback delivered by a trainer to a trainee in a live surgery. Extract the following aspects from given feedback line:\\
  action - identify any phrases that explicitly indicate an action the trainee is expected to perform. This may include instructions, commands, or recommendations. If no such actions are present, produce \"NONE\" \\
  Interpret unclear or incomplete feedback as concretely as possible, but avoid introducing assumptions not supported by the text. Interpret phases describing actions (e,g., \"you're gonna\", \"you will\", \"I want you to\", \"I like the\") as instructions for the trainee. \\
  Express extracted actions in the imperative form. Extract just the broad action with optional modifiers (e.g., 'come closer', 'tap litte', 'check left side', 'start doing'), but don't include the object of the action, such as anatomic component or instrument used. \\
  Produce output strictly in the format [action 1, action 2, ...].

  \medskip
  
  {\ttfamily
  Process this feedback line: \{feedback\_line\}
  }
  \end{tcolorbox}
\end{table*}

\begin{table*}[t]
  \captionsetup{aboveskip=4pt,belowskip=2pt} 
  \small
  \centering
  \caption{\textbf{Prompt for extracting \emph{tissue} component of triplet from a feedback line.}}
  \label{tab:tissue_extraction_prompt}
  \begin{tcolorbox}[
    enhanced, 
    width=\textwidth,
    colback=white,
    colframe=black!75!white,
    colbacktitle=black!85!white,
    coltitle=white,
    title=Tissue Extraction Prompt,
    fonttitle=\bfseries,
    left=3pt,right=3pt,top=3pt,bottom=3pt,boxsep=2pt,
    before skip=2pt, after skip=2pt
  ]
  \footnotesize

  \textbf{Instruction.} You are working in the context of verbal feedback delivered by a trainer to a trainee in a live robot-assystec daVinci surgery. Extract the following aspects from given feedback line:\\
  tissue, visual body part, or anatomic elements - identify any mentions of tissue, visual body part, or anatomic elements. If none is mentioned, produce \"NONE\" \\
  Make sure you extract all and every tissue, visual body part, or anatomic elements mentiond together with its property, if mentioned, including urether, bladder neck, small prostate, prostate contour, lateral thing, anteiror part or others. \\
  Some mentions are colloquial (e.g., bleeder is vein) or abbreviated (e.g., SV is Seminal Vesicles). Interpret these in the context of abdominal urologic surgery and express in proper medical form. \\
  If the term is ambiguous (e.g., \"lateral thing\" or \"anterior part\"), interpret it in the surgical context and include it as an approximate or general anatomical reference (e.g., \"lateral structure\" or \"anterior structure\"). \\
  Produce output strictly in the format [element 1, element 2, ...].

  \medskip
  
  {\ttfamily
  Process this feedback line: \{feedback\_line\}
  }
  \end{tcolorbox}
\end{table*}

\begin{table*}[h!]
    \captionsetup{aboveskip=4pt,belowskip=2pt}
    \small
    \centering
    \caption{\textbf{Prompt for Step 1: Initial fine-grained clustering of IAT mentions.}}
    \label{tab:initial_clustering_prompt}
    \begin{tcolorbox}[
        enhanced, 
        width=\textwidth,
        colback=white,
        colframe=black!75!white,
        colbacktitle=black!85!white,
        coltitle=white,
        title=Ontology Induction Prompt (Step 1: Fine-Grained Clustering),
        fonttitle=\bfseries,
        boxrule=0.4pt,
        arc=0pt, outer arc=0pt,
        left=3pt,right=3pt,top=3pt,bottom=3pt,boxsep=2pt,
        before skip=2pt, after skip=2pt
        ]
    \footnotesize

    \textbf{Role.} You are an expert in clinical language and surgical ontology design. Your task is to analyze a list of raw textual mentions extracted from surgeon feedback and group them into semantically coherent, fine-grained clusters.
    
    \medskip
    \textbf{Instructions}
    \begin{enumerate}[leftmargin=*,nosep,label=\arabic*.]
        \item You will be given a list of raw text mentions from a specific category: \textbf{Instrument}, \textbf{Action}, or \textbf{Tissue}.
        \item Group these mentions into clusters based on their semantic similarity and clinical meaning. It is better to create more specific, smaller clusters than to group dissimilar items.
            \item For each cluster, provide a concise and descriptive name that captures the core concept (e.g., "Energy Application", "Traction", "Vascular Structures").
        \item Handle synonyms, abbreviations, slight misspellings, and minor variations in phrasing by placing them in the same cluster.
    \end{enumerate}
    
    \textbf{Formatting}
    \begin{itemize}[leftmargin=*,nosep]
    \item Produce a structured list. For each cluster, first state the cluster name, then list all the member mentions.
    \item Do not include explanations or summaries outside of the structured list.
    \end{itemize}
    
    \textbf{Input}
    \begin{itemize}[leftmargin=*,nosep]
    \item \textit{Category} (str): The type of mentions provided (e.g., "Action").
    \item \textit{Mentions} (list of str): The raw textual mentions to be clustered.
    \end{itemize}
    
    \textbf{Output}
    \begin{itemize}[leftmargin=*,nosep]
    \item (str) A formatted string listing each cluster name followed by its members.
    \end{itemize}
    
    {\ttfamily
    Category: \{category\_name\}\\
    Mentions: \{list\_of\_mentions\}\\[2pt]
    Clustered Output:
    }
    \end{tcolorbox}
\end{table*}

\begin{table*}[h!]
    \captionsetup{aboveskip=4pt,belowskip=2pt}
    \small
    \centering
    \caption{\textbf{Prompt for Step 2: Merging fine-grained clusters into functional meta-clusters.}}
    \label{tab:merge_clustering_prompt}
    \begin{tcolorbox}[
    enhanced, 
    width=\textwidth,
    colback=white,
    colframe=black!75!white,
    colbacktitle=black!85!white,
    coltitle=white,
    title=Ontology Induction Prompt (Step 2: Merging into Meta-Clusters),
    fonttitle=\bfseries,
    boxrule=0.4pt,
    arc=0pt, outer arc=0pt,
    left=3pt,right=3pt,top=3pt,bottom=3pt,boxsep=2pt,
    before skip=2pt, after skip=2pt
    ]
    \footnotesize
    
    \textbf{Role.} You are a senior surgical educator and ontology designer refining a clinical label space. Your goal is to merge fine-grained clusters into broader, more practical "meta-clusters" that are functionally relevant for training a machine learning model.
    
    \medskip
    \textbf{Instructions}
    \begin{enumerate}[leftmargin=*,nosep,label=\arabic*.]
        \item You will receive a list of fine-grained clusters, where each cluster has a name and a list of members.
        \item Your task is to merge these fine-grained clusters into a smaller number of high-level meta-clusters based on their shared surgical \textbf{function} or \textbf{purpose}.
        \item For each new meta-cluster, assign a canonical, high-level name that is intuitive and clinically standard (e.g., coagulate, apply\_traction, prostate\_tissue).
        \item Use the provided few-shot examples to guide your merging logic. The goal is to create categories that are distinct and frequent enough for robust classification.
    \end{enumerate}
    
    \textbf{Formatting}
    \begin{itemize}[leftmargin=*,nosep]
    \item Produce a structured list. For each meta-cluster, state its new canonical name, followed by the names of the fine-grained clusters it now contains.
    \item Ensure the output is clean and directly usable as a mapping dictionary.
    \end{itemize}
    
    \textbf{Input}
    \begin{itemize}[leftmargin=*,nosep]
    \item \textit{Fine-Grained Clusters} (str): The structured output from the first clustering step.
    \item \textit{Merge Examples} (str): A few examples showing how to merge related clusters under a new meta-cluster name.
    \end{itemize}
    
    \textbf{Output}
    \begin{itemize}[leftmargin=*,nosep]
    \item (str) A formatted string showing the final meta-cluster mapping.
    \end{itemize}
    
    {\ttfamily
    Fine-Grained Clusters to Merge: \{structured\_list\_from\_step\_1\}\\
    Merge Examples: \{few\_shot\_examples\}\\[2pt]
    Merged Meta-Cluster Output:
    }
    \end{tcolorbox}
\end{table*}

\begin{table*}[t]
  \captionsetup{aboveskip=4pt,belowskip=2pt} 
  \small
  \centering
  \caption{\textbf{Prompt for scoring feedback generations and getting clinician-aligned fidelity via LLM-as-Judge.}}
  \label{tab:feedback_eval_prompt}
  \begin{tcolorbox}[
    enhanced, 
    width=\textwidth,
    colback=white,
    colframe=black!75!white,
    colbacktitle=black!85!white,
    coltitle=white,
    title=Generated Feedback Evaluation Prompt,
    fonttitle=\bfseries,
    left=3pt,right=3pt,top=3pt,bottom=3pt,boxsep=2pt,
    before skip=2pt, after skip=2pt
  ]
  \footnotesize

  \textbf{Instruction.} You are a surgical feedback evaluator. 
  
  Your goal is to compare the faithfulness of the generated feedback to the ground truth human expert feedback. Please judge it based only on the core action(s) the trainee is being asked to perform. This includes whether the feedback is a positive or negative (don’t/stop/avoid) instruction — these distinctions are important.

    \medskip
  Score the faithfulness on a scale from 1 to 5, defined as follows:
  
  1 - Opposite or unsafe action: Generated feedback communicates an action that is clearly opposite in intent, or unsafe/conflicting with the ground truth. Includes flipping a negative to a positive (or vice versa) when that changes meaning. (e.g., 'cut the vein' vs 'secure the vein', 'stop' vs 'continue', 'don’t dissect here' vs 'dissect here', 'only one clip' vs 'put all the clips').
  
  2 - Wrong or mismatched action: Generated feedback suggests an action that is different in type or modality from the ground truth. (e.g., 'sweep towards you' vs 'pull the tissue', 'buzz the artery' vs 'clip the artery', 'stop this bleeding' vs 'place a stitch').

    3 - Partially aligned, missing or adding key details: Generated feedback conveys the general action or intent (positive or negative), but omits or adds important details that change the amount, precision, target tissue/instrument, strength or safety of the instruction, or lacks explicit reference to the target action or tissue. (e.g., 'stop the bleed' vs 'buzz that bleeder', 'clip to the artery' vs 'only one clip to artery, wanna be safe', 'come closer' vs 'you can even come 1 mm closer to the prostate', 'do not do it' vs 'don’t do any blunt dissection').

    4 - Mostly aligned, minor wording/emphasis differences: Generated feedback matches the core action and target tissue/instrument, with only minor differences in degree of emphasis, polite hedging, phrasing or verbosity. (e.g., 'cauterize this' vs 'buzz that bleeder', 'closer to prostate' vs 'you can even come 1 mm closer to the prostate', 'clip the artery safely' vs 'only one clip to artery, wanna be safe').

    5 - Perfectly aligned: Generated feedback fully matches the ground truth in core action, target tissue/instrument, intent (positive or negative), and strength of instruction. (e.g., 'coag the vein' vs 'buzz that bleeder', 'stop this bleeding by cauterizing' vs 'buzz that bleeder', 'move the left hand under the ureter' vs 'get L hand below ureter').

    \medskip
    Using this scale, evaluate this generated feedback: "\{gen\_fb\}" against this ground truth feedback: "\{gt\_fb\}". Produce just the number, as your response needs to be processed automatically.

  \end{tcolorbox}
\end{table*}

\section{Task1: Video$\rightarrow$IAT Details}
\label{apx:task1_details}

\subsection{Confusion-Matrix Diagnostics for Video$\rightarrow$IAT}
\label{app:cm_iat}

\begin{figure*}[t]
  \centering
  \includegraphics[width=\textwidth,
                   height=0.33\textheight,
                   keepaspectratio]{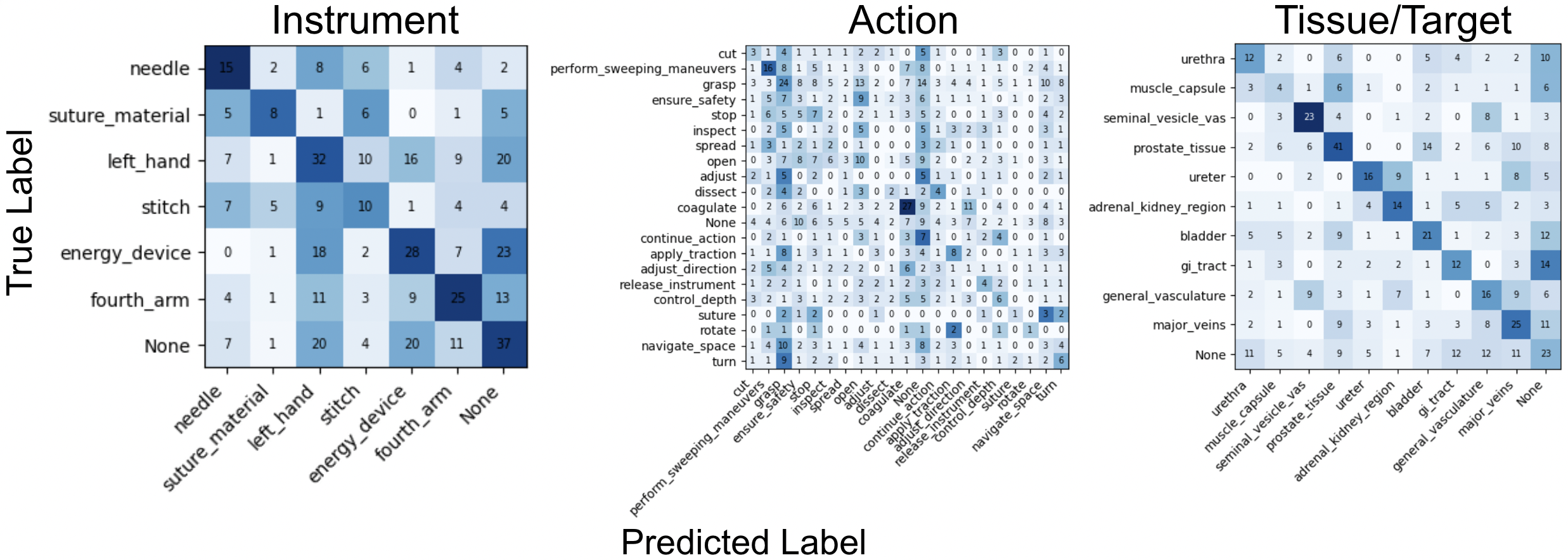}
  \vspace{-12pt}
  \caption{\textbf{Video$\rightarrow$IAT confusion matrices (combined).}
  Per-class confusion for the three prediction heads—Instrument, Action, and Tissue/Target—computed on the validation split and visualized side-by-side in a single panel. The plots highlight characteristic confusions (e.g., \emph{left\_hand} vs.\ \emph{fourth\_arm}, energy actions around \emph{coagulate}, and vascular classes such as \emph{general\_vasculature} vs.\ \emph{major\_veins}), as well as the impact of \emph{None} labels. These diagnostics complement the AUC results in Table~\ref{tab:video_iat_auc_gain_results} by illustrating where temporal tracking and clinical context reduce ambiguity.}
  \label{fig:cm_iat_combined}
  \vspace{-6pt}
\end{figure*}

Figure~\ref{fig:cm_iat_combined} shows combined confusion matrices for the
\emph{Instrument}, \emph{Action}, and \emph{Tissue/Target} heads on the validation
set. We observe expected, semantically proximate confusions—e.g., \emph{left\_hand}
vs.\ \emph{fourth\_arm}, energy-related verbs around \emph{coagulate}, and vascular
targets such as \emph{general\_vasculature} vs.\ \emph{major\_veins}—as well as
misclassifications into \emph{None}. These patterns align with the quantitative
gains in Table~\ref{tab:video_iat_auc_gain_results}, where adding clinical context
and temporal instrument tracking improves discrimination, particularly for
instrument and tissue components.

\begin{figure*}[ht!]
  \centering
  \includegraphics[width=\textwidth,
                   height=0.40\textheight,
                   keepaspectratio]{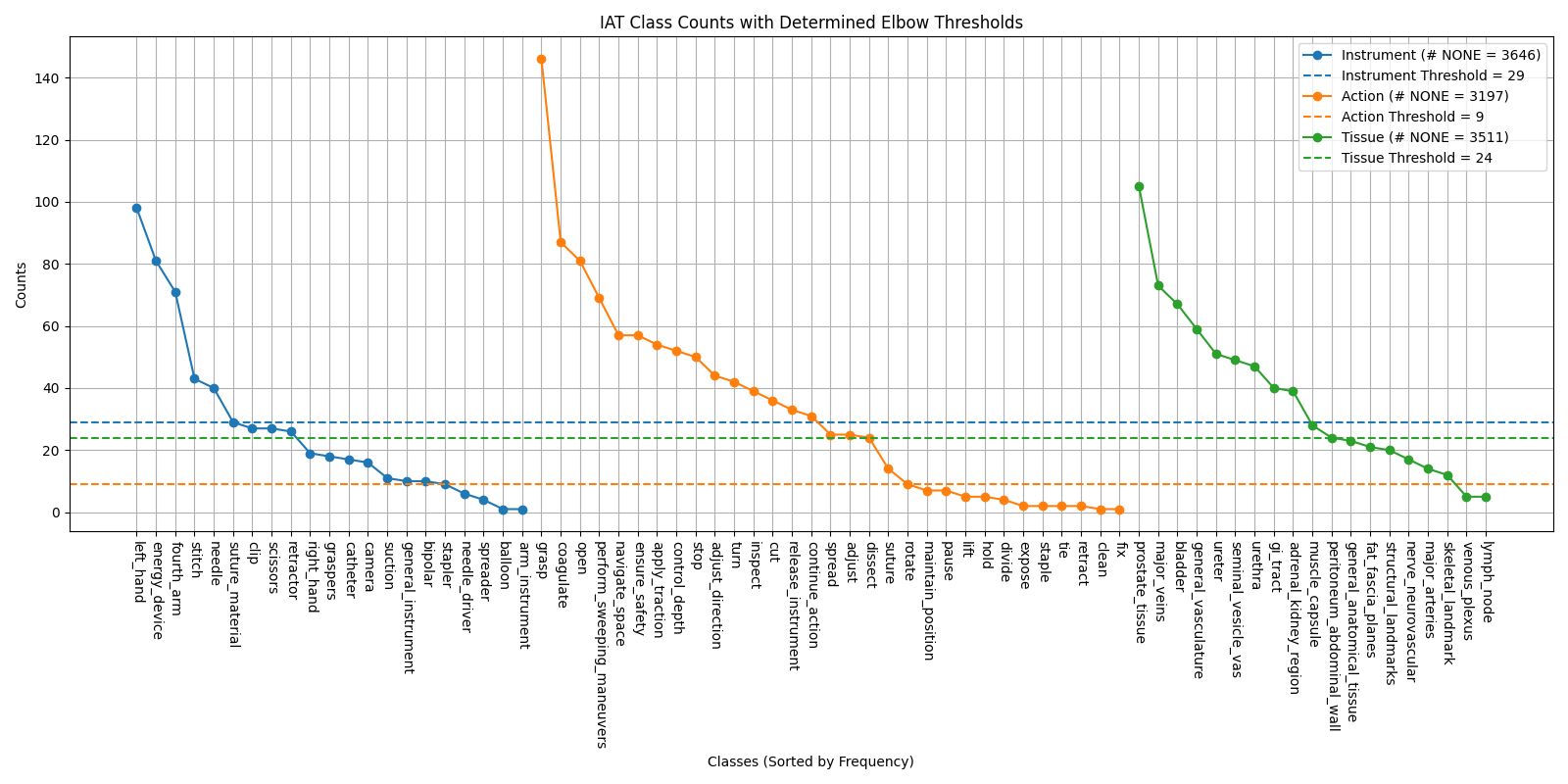}
  \vspace{-6pt}
  \caption{\textbf{IAT class frequency and elbow thresholds.} Class counts for instruments, actions, and tissues (sorted by frequency). Dashed lines indicate elbow‐derived cutoffs (29/9/24); legend also reports the number of feedback lines with \texttt{NONE} for that triplet component.}
  \label{fig:iat_class_counts_with_thresholds}
  \vspace{-6pt}
\end{figure*}

\section{Training and Implementation Details}
\label{apd:training_details}

This section provides detailed information on the training procedures and model configurations used in our experiments.

\parahead{Task 1: Video $\rightarrow$ IAT Prediction.}
The models for this task are trained and evaluated using a 5-fold stratified cross-validation scheme. Each of the three heads (Instrument, Action, Tissue) is trained as an independent multi-class classifier.
\begin{itemize}[leftmargin=*, itemsep=-0.3em, topsep=2pt]
    \item \textbf{Context-Only Models:} For baselines that fuse video and clinical context embeddings without temporal tracking, we use an `MLPClassifier` from Scikit-learn. The network consists of three hidden layers with (64, 32, 16) units respectively, using a ReLU activation function and the Adam optimizer, trained for a maximum of 1000 iterations.
   \item \textbf{Temporal Tracking Model (Ours):} Our full model uses a two-stage hybrid approach to incorporate motion. \textbf{Stage 1 (Learning Motion Representation):} We first train a PyTorch fusion model containing an LSTM component. This initial end-to-end training teaches the LSTM to extract task-relevant features from instrument coordinate sequences. The model's core is a 10-layer LSTM with 32 hidden units, and it is trained for 20 epochs (Adam optimizer, LR $1 \times 10^{-3}$, batch size 8). \textbf{Stage 2 (Final Classification):} After Stage 1, the trained LSTM component is used as a fixed feature extractor to generate a motion embedding for each video clip. These motion embeddings are then concatenated with the video and clinical context embeddings. A Scikit-learn `MLPClassifier` is trained on this final combined feature vector to perform the classification. This MLP's architecture is identical to our context-only baselines: three hidden layers (64, 32, 16) with ReLU activation, trained for up to 1000 iterations.
\end{itemize}

\parahead{Task 2: Feedback Generation.} All fine-tuned models are trained for three full passes over the training data unless otherwise specified.
\begin{itemize}[leftmargin=*, itemsep=-0.3em, topsep=2pt]
    \item \textbf{VQA LLaMA (11B):} This baseline uses a pre-trained `unsloth/llama-3.2-11b-vision-instruct-bnb-4bit` model. We frame the task as visual question answering, prompting the model with the procedure, task, and a single video frame to generate the corresponding feedback. The model is fine-tuned using Low-Rank Adaptation (LoRA, rank=16) on all self-attention projection layers (`q\_proj`, `k\_proj`, `v\_proj`, `o\_proj`, etc.). From each 10-second video clip, we create 10 separate training instances by uniformly sampling 10 frames and pairing each frame individually with the ground-truth feedback. Training is performed using an 8-bit AdamW optimizer with a learning rate of $2 \times 10^{-4}$ and a linear scheduler.
    \item \textbf{SurgVLP Gen:} This baseline is a video captioning model composed of a SurgVLP encoder (ResNet-50 backbone) and a standard GPT-2 decoder. A trainable two-layer mapping network translates the video embedding into a soft prompt (prefix length of 8 tokens) for the decoder. The video embedding is derived by mean-pooling the features of 3 uniformly sampled frames from each clip. The entire model—encoder, mapper, and decoder—is fine-tuned end-to-end for 30 epochs using an AdamW optimizer with distinct learning rates for the mapper ($3 \times 10^{-4}$) and the language model ($5 \times 10^{-6}$), with a cosine learning rate schedule.
\end{itemize}

\section{Additional Results for IAT Triplet Extraction}
\label{apd:additional_metrics}
Tables \ref{tab:prec_rec_IAT} and \ref{tab:f1_score_IAT} report additional metrics for the IAT prediction task from video using SurgVLP, HecVL, and PeskaVLP surgery specific models. Table~\ref{tab:video_iat_auc_additional_results} complements the main-text Table~\ref{tab:video_iat_auc_gain_results} by presenting AUCs for additional video encoders evaluated under \emph{full context} (Vision + Procedure/Task + Temporal Instrument Tracking). As an anchor, a generic, non-surgical LLM without a specialized video encoder or fine-tuning (GPT-4o) operates near chance on AUC ($\approx 0.5$), whereas self-/pre-trained video encoders (VideoMAE, V-JEPA-2) yield clear gains, particularly for \textit{Instrument} and \textit{Tissue}. \textit{Action} remains the hardest subtask across models.

\paragraph{Model and implementation notes.}
Beyond SurgVLP and PeskaVLP (evaluated in stages: Vision $\rightarrow$ +Procedure $\rightarrow$ +Task $\rightarrow$ +Temporal Tracking), we include strong encoder baselines under identical full-context inputs. Specifically, we evaluate \emph{VideoMAE} (Kinetics-400 pretrained\footnote{\url{https://huggingface.co/MCG-NJU/videomae-base}})~\citep{tong2022videomae} and a \emph{surgery-pretrained VideoMAE\footnote{\url{https://github.com/arushig100/Multi-Modal-SSL-for-Surgical-Feedback-Effectiveness-Assessment}}} from prior ML4H work~\citep{gupta2025multi}, as well as \emph{V-JEPA-2} (ViT-L, fpc64-256, base checkpoint\footnote{\url{https://huggingface.co/facebook/vjepa2-vitl-fpc64-256}})~\citep{assran2025v}. For non–GPT-4o rows, text features use the MedEmbed sentence transformer\footnote{ \url{https://github.com/abhinand5/MedEmbed}.}. To our knowledge there is no publicly released surgery-finetuned V-JEPA-2; we therefore report its base encoder.

\paragraph{Interpretation.}
Compared to GPT-4o without a dedicated video backbone, self-/pre-trained video encoders substantially improve AUC, with surgery-pretrained VideoMAE strongest on \textit{Tissue} (0.77) and V-JEPA-2 slightly ahead on \textit{Instrument} (0.69). These trends align with the main-text finding that adding clinically meaningful context and temporal instrument tracking is most beneficial for \textit{Instrument} and \textit{Tissue}, while \textit{Action} remains more challenging.

\begin{table*}[h!]
\small
\centering

\caption{Precision and Recall scores for IAT triplet prediction from video task.}
\label{tab:prec_rec_IAT}

\begin{tabular}{llcccccc}
\toprule
\textbf{Base Model} & \textbf{Context} & 
\multicolumn{2}{c}{\textbf{Instrument}} & 
\multicolumn{2}{c}{\textbf{Action}} & 
\multicolumn{2}{c}{\textbf{Tissue}} \\
\cmidrule(lr){3-4} \cmidrule(lr){5-6} \cmidrule(lr){7-8}
& & Precision & Recall & Precision & Recall & Precision & Recall \\
\midrule
\multirow{4}{*}{SurgVLP} 
& Vision & $0.29_{\pm.03}$ & $0.29_{\pm.02}$ & $0.09_{\pm.01}$ & $0.09_{\pm.01}$ & $0.34_{\pm.03}$ & $0.29_{\pm.01}$ \\
& +Procedure & $0.31_{\pm.03}$ & $0.31_{\pm.02}$ & $0.11_{\pm.02}$ & $0.10_{\pm.02}$ & $0.32_{\pm.04}$ & $0.29_{\pm.03}$ \\
& +Task & $0.32_{\pm.05}$ & $0.31_{\pm.04}$ & {\boldmath $0.13_{\pm.04}$} & {\boldmath $0.12_{\pm.03}$} & {\boldmath $0.35_{\pm.04}$} & $0.33_{\pm.04}$ \\
& +Temporal Tracking & {\boldmath $0.33_{\pm.04}$} & {\boldmath $0.35_{\pm.05}$} & $0.11_{\pm.02}$ & $0.11_{\pm.02}$ & {\boldmath $0.35_{\pm.01}$} & {\boldmath $0.35_{\pm.02}$} \\
\midrule

\multirow{4}{*}{HecVL} 
& Vision & $0.35_{\pm.06}$ & $0.35_{\pm.05}$ & $0.11_{\pm.02}$ & $0.11_{\pm.02}$ & $0.32_{\pm.05}$ & $0.31_{\pm.04}$ \\
& +Procedure & $0.30_{\pm.05}$ & $0.30_{\pm.05}$ & $0.11_{\pm.01}$ & {\boldmath $0.12_{\pm.01}$} & $0.34_{\pm.03}$ & $0.32_{\pm.02}$ \\
& +Task & $0.33_{\pm.07}$ & $0.34_{\pm.07}$ & $0.10_{\pm.02}$ & $0.11_{\pm.02}$ & $0.34_{\pm.04}$ & $0.33_{\pm.04}$ \\
& +Temporal Tracking & {\boldmath $0.39_{\pm.07}$} & {\boldmath $0.39_{\pm.07}$} & {\boldmath $0.12_{\pm.02}$} & {\boldmath $0.12_{\pm.01}$} & {\boldmath $0.38_{\pm.04}$} & {\boldmath $0.36_{\pm.03}$} \\
\midrule

\multirow{4}{*}{PeskaVLP} 
& Vision & $0.33_{\pm.06}$ & $0.32_{\pm.03}$ & $0.10_{\pm.04}$ & $0.10_{\pm.03}$ & $0.32_{\pm.04}$ & $0.32_{\pm.04}$ \\
& +Procedure & $0.30_{\pm.05}$ & $0.29_{\pm.05}$ & $0.09_{\pm.02}$ & $0.10_{\pm.02}$ & $0.32_{\pm.04}$ & $0.30_{\pm.04}$ \\
& +Task & $0.37_{\pm.04}$ & $0.34_{\pm.05}$ & $0.12_{\pm.02}$ & $0.11_{\pm.02}$ & $0.35_{\pm.04}$ & $0.33_{\pm.04}$ \\
& +Temporal Tracking & {\boldmath $0.39_{\pm.05}$} & {\boldmath $0.37_{\pm.04}$} & {\boldmath $0.13_{\pm.01}$} & {\boldmath $0.13_{\pm.02}$} & {\boldmath $0.41_{\pm.04}$} & {\boldmath $0.40_{\pm.03}$} \\
\bottomrule
\end{tabular}
\end{table*}

\begin{table*}[h!]
\small
\centering

\caption{\textbf{F1 scores for IAT triplet prediction from video.} Macro-averaged F1 and relative gains for Instrument (7-way), Action (21-way), and Tissue (11-way) prediction across context inputs. A uniform random top-1 guess would achieve only about $1/C$ F1 (roughly 0.14, 0.05, and 0.09 for Instrument, Action, and Tissue, respectively). Our best models reach F1 $\approx 0.33$--$0.39$ for Instrument and $\approx 0.34$--$0.39$ for Tissue (around 2--4$\times$ chance), while Action remains the hardest 21-way task with F1 $\approx 0.10$--$0.12$ (about 2--3$\times$ chance), and still benefits consistently from added context and temporal tracking.}

\label{tab:f1_score_IAT}

\begin{tabular}{llcccccc}
\toprule
\textbf{Base Model} & \textbf{Context} & 
\multicolumn{2}{c}{\textbf{Instrument}} & 
\multicolumn{2}{c}{\textbf{Action}} & 
\multicolumn{2}{c}{\textbf{Tissue}} \\
\cmidrule(lr){3-4} \cmidrule(lr){5-6} \cmidrule(lr){7-8}
& & F1 & Gain & F1 & Gain & F1 & Gain \\
\midrule
\multirow{4}{*}{SurgVLP} 
& Vision & $0.28_{\pm0.02}$ & - & $0.09_{\pm0.01}$ & - & $0.30_{\pm0.01}$ & - \\
& + Procedure & $0.29_{\pm0.02}$ & \uab{3.6\%} & $0.10_{\pm0.02}$ & \uab{11.1\%} & $0.29_{\pm0.03}$ & \dar{3.3\%} \\
& + Task & $0.31_{\pm0.05}$ & \uab{10.7\%} & {\boldmath $0.12_{\pm0.03}$} & \uab{33.3\%} & $0.33_{\pm0.04}$ & \uab{10.0\%} \\
& + Temporal Tracking (our) & {\boldmath $0.33_{\pm0.04}$} & \uab{17.9\%} & $0.10_{\pm0.02}$ & \uab{11.1\%} & {\boldmath $0.34_{\pm0.02}$} & \uab{13.3\%} \\

\midrule
\multirow{4}{*}{HecVL} 
& Vision & $0.28_{\pm0.05}$ & - & $0.10_{\pm0.01}$ & - & $0.30_{\pm0.04}$ & - \\
& + Procedure & $0.29_{\pm0.04}$ & \uab{3.6\%} & $0.11_{\pm0.01}$ & \uab{10.0\%} & $0.32_{\pm0.03}$ & \uab{6.7\%} \\
& + Task & $0.33_{\pm0.06}$ & \uab{17.9\%} & $0.10_{\pm0.02}$ & \dar{0.0\%} & $0.32_{\pm0.04}$ & \uab{6.7\%} \\
& + Temporal Tracking (our) & {\boldmath $0.38_{\pm0.07}$} & \uab{35.7\%} & {\boldmath $0.12_{\pm0.01}$} & \uab{20.0\%} & {\boldmath $0.36_{\pm0.04}$} & \uab{20.0\%} \\

\midrule
\multirow{4}{*}{PeskaVLP} 
& Vision & $0.32_{\pm0.03}$ & - & $0.10_{\pm0.03}$ & - & $0.32_{\pm0.04}$ & - \\
& + Procedure & $0.30_{\pm0.05}$ & \dar{6.3\%} & $0.10_{\pm0.02}$ & \dar{0.0\%} & $0.30_{\pm0.04}$ & \dar{6.3\%} \\
& + Task & $0.34_{\pm0.04}$ & \uab{6.3\%} & $0.11_{\pm0.02}$ & \uab{10.0\%} & $0.33_{\pm0.04}$ & \uab{3.1\%} \\
& + Temporal Tracking (our) & {\boldmath $0.36_{\pm0.03}$} & \uab{12.5\%} & {\boldmath $0.12_{\pm0.01}$} & \uab{20.0\%} & {\boldmath $0.39_{\pm0.03}$} & \uab{21.9\%} \\
\bottomrule
\end{tabular}
\end{table*}

\begin{table*}[ht!]
\small
\centering

\caption{\textbf{IAT triplet prediction from video using additional models.} All reported scores are AUCs for models evaluated using full context (Vision + Procedure/Task + Temporal Tracking). These results are complementary to the ones reported in Table \ref{tab:video_iat_auc_gain_results}}

\vspace{-7.0pt}
\label{tab:video_iat_auc_additional_results}

\begin{tabular}{lccc}
\toprule
\textbf{Base Model} & 
\textbf{Instrument} & 
\textbf{Action} & 
\textbf{Tissue} \\

\midrule
GPT-4o & $0.50_{\pm0.01}$ & $0.51_{\pm0.01}$ & $0.52_{\pm0.02}$ \\
VideoMAE (\href{https://huggingface.co/MCG-NJU/videomae-base}{kinetics-pretrained}) \citep{tong2022videomae} & $0.68_{\pm0.02}$ & $0.58_{\pm0.02}$ & $0.74_{\pm0.01}$ \\
VideoMAE (\href{https://github.com/arushig100/Multi-Modal-SSL-for-Surgical-Feedback-Effectiveness-Assessment}{surgery-pretrained}) \citep{gupta2025multi} & $0.65_{\pm0.02}$ & $0.60_{\pm0.02}$ & $0.77_{\pm0.02}$ \\
V-JEPA-2 (\href{https://huggingface.co/facebook/vjepa2-vitl-fpc64-256}{base}) \citep{assran2025v} & $0.69_{\pm0.02}$ & $0.57_{\pm0.02}$ & $0.73_{\pm0.01}$ \\

\bottomrule
\end{tabular}

\end{table*}

\begin{table*}[ht!]
\small
\centering

\caption{\textbf{Performance of automated Surgical Action Triplet recognition from surgical video.} Reporting AUCs and relative gains for Instrument, Action, and Tissue prediction across different context inputs with SurgVLP, HecVL, and PeskaVLP base models. We observe consistent gains from added clinical context and further improvements from our \emph{\textbf{temporal instrument tracking}}. Superscripts mark statistically significant AUC gains compared to the previous context condition (paired DeLong test with Holm correction within base model: $^{**}p_{\text{adj}}<0.01$, $^{*}p_{\text{adj}}<0.05$); analysis details in Appendix \ref{apd:stat_tests_table2}.}
\vspace{-7.0pt}
\label{tab:video_iat_auc_gain_results}

\begin{tabular}{llcccccc}
\toprule
\textbf{Base Model} & \textbf{Context} & 
\multicolumn{2}{c}{\textbf{Instrument}} & 
\multicolumn{2}{c}{\textbf{Action}} & 
\multicolumn{2}{c}{\textbf{Tissue}} \\
\cmidrule(lr){3-4} \cmidrule(lr){5-6} \cmidrule(lr){7-8}
& & AUC & Gain & AUC & Gain & AUC & Gain \\

\midrule
GPT-4o & Vision+Proc./Task+Tracking & $0.50_{\pm0.01}$ & - & $0.51_{\pm0.01}$ & - & $0.52_{\pm0.02}$ & - \\
VideoMAE (kinetics) & Vision+Proc./Task+Tracking & $0.68_{\pm0.02}$ & - & $0.58_{\pm0.02}$ & - & $0.74_{\pm0.01}$ & - \\
VideoMAE (surgery) & Vision+Proc./Task+Tracking & $0.65_{\pm0.02}$ & - & $0.60_{\pm0.02}$ & - & $0.77_{\pm0.02}$ & - \\
V-JEPA-2 (base) & Vision+Proc./Task+Tracking & $0.69_{\pm0.02}$ & - & $0.57_{\pm0.02}$ & - & $0.73_{\pm0.01}$ & - \\

\midrule
\multirow{4}{*}{SurgVLP} 
& Vision & $0.65_{\pm0.02}$ & - & $0.58_{\pm0.02}$ & - & $0.70_{\pm0.01}$ & - \\
& + Procedure & $0.67_{\pm0.02}$ & \uab{3.1\%} & $0.58_{\pm0.02}$ & \dar{0.0\%} & $0.73_{\pm0.01}$ & \uab{4.3\%} \\
& + Task & $0.70_{\pm0.01}$ & \uab{7.7\%} & $0.60^{*}_{\pm0.01}$ & \uab{3.4\%}  & $0.76_{\pm0.02}$ & \uab{8.6\%} \\
& + Temporal Tracking (our) & {\boldmath $0.73_{\pm0.03}$} & \uab{12.3\%} & {\boldmath $0.61_{\pm0.01}$} & \uab{5.2\%} & {\boldmath $0.79^{**}_{\pm0.01}$} & \uab{12.9\%} \\

\midrule
\multirow{4}{*}{HecVL} 
& Vision & $0.68_{\pm0.04}$ & - & $0.60_{\pm0.01}$ & - & $0.74_{\pm0.02}$ & - \\
& + Procedure & $0.68_{\pm0.03}$ & \dar{0.0\%} & $0.60_{\pm0.01}$ & \dar{0.0\%} & $0.76_{\pm0.02}$ & \uab{2.7\%} \\
& + Task & $0.71_{\pm0.04}$ & \uab{4.4\%} & $0.61_{\pm0.01}$ & \uab{1.7\%}  & $0.77_{\pm0.03}$ & \uab{4.1\%} \\
& + Temporal Tracking (our) & {\boldmath $0.74_{\pm0.03}$} & \uab{8.8\%} & {\boldmath $0.62_{\pm0.01}$} & \uab{3.3\%} & {\boldmath $0.77_{\pm0.01}$} & \uab{4.1\%} \\

\midrule    
\multirow{4}{*}{PeskaVLP} 
& Vision & $0.67_{\pm0.03}$ & - & $0.60_{\pm0.02}$ & - & $0.74_{\pm0.02}$ & - \\
& + Procedure & $0.68_{\pm0.03}$ & \uab{1.5\%} & $0.59_{\pm0.02}$ & \dar{1.7\%} & $0.75_{\pm0.02}$ & \uab{1.4\%} \\
& + Task & $0.69_{\pm0.03}$ & \uab{3.0\%} & $0.60_{\pm0.01}$ & \dar{0.0\%} & $0.74_{\pm0.02}$ & \dar{0.0\%} \\
& + Temporal Tracking (our) & {\boldmath $0.74^{*}_{\pm0.03}$} & \uab{10.4\%} & {\boldmath $0.63_{\pm0.02}$} & \uab{5.0\%} & {\boldmath $0.79^{*}_{\pm0.01}$} & \uab{6.8\%} \\

\bottomrule
\end{tabular}

\end{table*}


\section{Details of Statistical Analysis and Results}

\subsection{Statistical Tests for Table~\ref{tab:video_iat_auc_gain_results}}
\label{apd:stat_tests_table2}

Our primary evaluation metric for Task~1 (Video$\rightarrow$IAT prediction) is the area under the ROC curve (AUC). To assess whether successive additions of context and temporal tracking yield statistically reliable improvements, we performed paired comparisons of AUC between adjacent conditions for each base model and task. For each base model (SurgVLP, HecVL, PeskaVLP) and each addition to the context (Vision $\rightarrow$ +Procedure, +Procedure $\rightarrow$ +Task, +Task $\rightarrow$ +Temporal Tracking), we:

\begin{enumerate}[leftmargin=*, itemsep=-0.3em, topsep=2pt]
    \item Computed class-level ROC curves and AUCs for each of the three IAT components (Instrument, Action, Tissue) on the same held-out examples under both conditions.
    \item Applied the paired DeLong test to obtain a $z$-score and $p$-value for the difference in AUC for each component.\footnote{We follow the standard implementation of the DeLong test for correlated ROC curves.}
    \item Aggregated the three component-level $z$-scores into a single task-level statistic using a weighted Stouffer method, with weights proportional to the number of positive examples per component. This yields one combined $z$-score and $p$-value per task and upgrade step.
    \item Controlled for multiple comparisons \emph{within} each base model (9 tests = 3 tasks $\times$ 3 steps) using the Holm procedure, yielding adjusted $p_{\text{adj}}$ values.
\end{enumerate}

We treat $p_{\text{adj}} \leq 0.05$ as statistically significant and report these effects in Table~\ref{tab:video_iat_auc_gain_results} using superscripts (** $p_{\text{adj}}<0.01$, * $p_{\text{adj}}<0.05$). For completeness, we also reserve \textsuperscript{\textdagger} to denote exploratory effects at $p \leq 0.10$ (uncorrected), though no such effects are highlighted in Table~\ref{tab:video_iat_auc_gain_results}.

The following upgrade steps showed statistically significant gains in AUC:

\begin{itemize}[leftmargin=*, itemsep=-0.3em, topsep=2pt]
    \item \textbf{SurgVLP, Action}: +Task vs.\ +Procedure, $\Delta\text{AUC} = +0.02$ (0.58 $\rightarrow$ 0.60), combined DeLong $p = 0.0051$, Holm-adjusted $p_{\text{adj}} = 0.041$ (marked with * in the Action gain for the +Task row).
    \item \textbf{SurgVLP, Tissue}: +Temporal Tracking vs.\ +Task, $\Delta\text{AUC} = +0.03$ (0.76 $\rightarrow$ 0.79), combined DeLong $p = 2.9\times 10^{-5}$, Holm-adjusted $p_{\text{adj}} = 2.6\times 10^{-4}$ (marked with ** in the Tissue gain for the +Temporal Tracking row).
    \item \textbf{PeskaVLP, Instrument}: +Temporal Tracking vs.\ +Task, $\Delta\text{AUC} = +0.05$ (0.69 $\rightarrow$ 0.74), combined DeLong $p = 0.0072$, Holm-adjusted $p_{\text{adj}} = 0.046$ (marked with * in the Instrument gain for the +Temporal Tracking row).
    \item \textbf{PeskaVLP, Tissue}: +Temporal Tracking vs.\ +Task, $\Delta\text{AUC} = +0.05$ (0.74 $\rightarrow$ 0.79), combined DeLong $p = 0.0016$, Holm-adjusted $p_{\text{adj}} = 0.014$ (marked with * in the Tissue gain for the +Temporal Tracking row).
\end{itemize}

These results substantiate that (i) task-level context provides a statistically reliable improvement for the Action component under SurgVLP, and (ii) temporal tracking produces robust gains for Instrument and Tissue under both backbones where the largest AUC increases are observed, complementing the descriptive improvements reported in Table~\ref{tab:video_iat_auc_gain_results}.

\subsection{Statistical Tests for Table~\ref{tab:alignment_text_scores}}
\label{apd:stat_tests_table3}

We compared ordinal \textit{Fidelity} scores (1--5) across the GPT-4o variants in Table~\ref{tab:alignment_text_scores} against the strongest baseline (GPT-4o, Video+context). Because procedure and task may affect difficulty, the primary analysis used a stratified Wilcoxon (van Elteren) test by procedure$\times$task. As a robustness check, we also ran pooled Mann--Whitney U tests (Wilcoxon rank-sum) with Holm family-wise error correction.

For effect sizes, we report Vargha--Delaney $A_{12}$ (probability that a random item from group A, the baseline, scores higher than a random item from group B, the variant) and Cliff's $\delta$ (rank-based effect size in $[-1,1]$). We interpret $|\delta|$ based on \cite{romano2006appropriate}: $\approx 0.00$--$0.147$ small, $0.147$--$0.33$ medium, $>0.33$ large.

Let \textbf{Baseline} denote GPT-4o, Video+context ($n = 1405$, mean $\approx 2.17$), we compare:

\begin{itemize}[leftmargin=*, itemsep=-0.3em, topsep=2pt]
    \item \textbf{IAT-only (GPT-4o +(I,A,T)) vs.\ Baseline} ($n = 1449$, mean $2.253$):\\
    Mann--Whitney $U = 976{,}413$, Holm-adjusted $p = 2.69\times 10^{-5}$ ($p < 0.001$).\\
    Effect sizes: $A_{12} = 0.465$, Cliff's $\delta = -0.071$. Interpreting $A_{12}$ relative to the variant, the IAT-only variant wins in about $1 - A_{12} \approx 0.535$ (53.5\%) of random baseline--variant pairings, a small effect.
    
    \item \textbf{Context+IAT (GPT-4o Context+(I,A,T)) vs.\ Baseline} ($n = 1364$, mean $2.326$):\\
    Mann--Whitney $U = 858{,}247$, Holm-adjusted $p = 1.06\times 10^{-13}$ ($p < 10^{-12}$).\\
    Effect sizes: $A_{12} = 0.434$, Cliff's $\delta = -0.132$. The variant wins in about $1 - A_{12} \approx 0.566$ (56.6\%) of random pairings, still a small effect but larger than IAT-only.
    
    \item \textbf{Context+IAT+gate (GPT-4o +confidence gate) vs.\ Baseline} ($n = 228$, mean $2.443$):\\
    Mann--Whitney $U = 128{,}584$, Holm-adjusted $p = 7.36\times 10^{-12}$ ($p < 10^{-11}$).\\
    Effect sizes: $A_{12} = 0.389$, Cliff's $\delta = -0.222$. The gated variant wins in about $1 - A_{12} \approx 0.611$ (61.1\%) of random pairings, corresponding to a small-to-moderate effect.
\end{itemize}

In summary, compared to the GPT-4o Video+context baseline, each added layer of structure yields a statistically significant improvement in ordinal \textit{Fidelity} scores after Holm correction. IAT-only provides a small uplift; Context+IAT yields a stronger small uplift; and Context+IAT+gate achieves a small-to-moderate effect, consistent with the monotonic gains in mean Alignment Score reported in Table~\ref{tab:alignment_text_scores}.

\section{Task 2: Feedback Generation Details}
\label{app:task2_details}

\noindent\textbf{Qualitative examples.}
Table~\ref{tab:iat_feedback_examples} illustrates representative cases from \emph{Task 2: Feedback Generation}. For each procedure/task, we list the trainer’s ground-truth feedback, the predicted \emph{Instrument–Action–Tissue (IAT)} triplet used to condition GPT-4o (prompt in App.~\ref{app:gpt4o_fbk_gen_prompt}), the model’s generated message, and the LLM-as-judge \emph{Alignment} score (1–5). These examples show how triplet grounding steers outputs toward specific, clinically meaningful guidance, and where residual gaps remain.

\newcommand{\iat}[3]{%
  \begin{tabular}[t]{@{}l@{}}%
  \texttt{inst: #1}\\
  \texttt{act:\ #2}\\
  \texttt{targ:\ #3}%
  \end{tabular}%
}

\begin{table*}[t]
\centering
\small
\setlength{\tabcolsep}{6pt}
\renewcommand{\arraystretch}{1.2}
\begin{tabular}{%
  >{\raggedright\arraybackslash}p{0.16\textwidth}%
  >{\raggedright\arraybackslash}p{0.16\textwidth}%
  >{\raggedright\arraybackslash}p{0.20\textwidth}%
  >{\raggedright\arraybackslash}p{0.30\textwidth}%
  >{\centering\arraybackslash}p{0.05\textwidth}%
}
\toprule
\textbf{Procedure/Task} & \textbf{Ground Truth Feedback} & \textbf{IAT triplet} & \textbf{Generated Feedback} & \textbf{Score} \\
\midrule

\addlinespace
Radical prostatectomy / Lymph node dissection &
just \textbf{open up}\ldots &
\iat{--}{open}{--} &
Don't hesitate to \textbf{enlarge the incision} here; you won't cause significant issues, so \textbf{ensure you have adequate visibility}. &
5 \\
\addlinespace
Radical prostatectomy / Bladder Neck &
be careful, don't \textbf{rip} the \textbf{prostate} &
\iat{--}{--}{prostate-tissue} &
Be cautious to avoid inadvertently entering the \textbf{prostate tissue}; maintain awareness of surrounding structures. &
5 \\

Simple prostatectomy / Adenoma dissection &
just remember to \textbf{stay right on the surface} of the \textbf{adenoma} &
\iat{--}{stay right}{prostate-tissue} &
Ensure you maintain \textbf{close proximity} to the \textbf{prostate tissue} to facilitate precise dissection and minimize trauma to surrounding structures. &
4 \\
\addlinespace
Partial nephrectomy / Bowel mobilization &
\textbf{lift up} with your \textbf{left hand} more &
\iat{left-hand}{lift}{--} &
Ensure your \textbf{left hand} is positioned to \textbf{provide optimal retraction and exposure}, enhancing visibility and access during bowel mobilization. &
4 \\
\addlinespace
Radical prostatectomy / Lymph node dissection &
so be real careful until you see your \textbf{ureter}, what you're \textbf{buzzing} &
\iat{--}{coagulate}{ureter} &
Avoid \textbf{coagulating} the \textbf{ureter} as it is a critical structure that should be preserved; ensure you correctly identify and protect it. &
4 \\

\addlinespace
Radical Prostatectomy / Dropping Bladder &
grab here, right here. &
\iat{--}{grasp}{--} &
Ensure your \textbf{grasp} is centered and balanced to maintain control and avoid unnecessary tension on the surrounding tissues during bladder mobilization. &
3 \\

\addlinespace
Simple Prostatectomy / Adenoma Dissection &
scrape it off &
\iat{--}{sweeping}{--} &
When performing \textbf{sweeping maneuvers}, ensure you use the side of the instrument effectively to clear loose tissue and define anatomical planes. &
3 \\

\addlinespace
Simple prostatectomy / Closing Bladder &
No grab closer to the needle. &
\iat{--}{grasp}{needle} &
Ensure you securely \textbf{grasp} the \textbf{needle} at its midpoint to maintain control and precision during suturing, avoiding unnecessary movement that could compromise the bladder closure. &
1 \\
\addlinespace
Radical prostatectomy / Lymph Node Dissection &
wait wait, dont buzz until they paralyze him &
\iat{--}{coagulate}{--} &
Ensure precise application of energy during \textbf{coagulation} to maintain hemostasis without inadvertently affecting surrounding tissues, especially near critical structures. &
1 \\

\bottomrule
\end{tabular}
\caption{Examples of surgeon-aligned feedback generation. For each procedure/task, we show the trainer’s ground-truth feedback, the predicted Instrument–Action–Target (IAT) triplet, the model’s generated feedback, and the alignment score (1–5).}
\label{tab:iat_feedback_examples}
\end{table*}

\section{Procedure Definitions Used for Clinical Context}
\label{app:proc_defs}
To condition the text encoder with situational priors (Sec.~\ref{subsec:clinical_context}), we expanded procedure names into concise, clinically phrased summaries using GPT\mbox{-}4o. Table~\ref{tab:proc_definitions} lists the definitions used during inference.

\begin{table*}[t]
\centering
\small
\setlength{\tabcolsep}{8pt}
\renewcommand{\arraystretch}{1.15}
\begin{tabular}{%
    >{\raggedright\arraybackslash}p{0.20\textwidth} 
    >{\raggedright\arraybackslash}p{0.05\textwidth} 
    >{\raggedright\arraybackslash}p{0.72\textwidth} 
}
\toprule
\textbf{Procedure} & \textbf{Count} & \textbf{Definition (GPT-4o-generated)} \\
\midrule
radical prostatectomy & 822 & Radical prostatectomy is a surgical procedure that involves the complete removal of the prostate gland along with some surrounding tissue, including the seminal vesicles and sometimes nearby lymph nodes. This procedure is primarily performed to treat localized prostate cancer and aims to eliminate cancerous cells while preserving as much surrounding healthy tissue as possible. It can be done through open surgery or minimally invasive techniques, such as laparoscopic or robotic-assisted surgery. Postoperative recovery may involve managing urinary incontinence and erectile dysfunction, which are common side effects. \\ 
simple prostatectomy & 279 & A simple prostatectomy is a surgical procedure aimed at removing the prostate gland and some surrounding tissue to alleviate symptoms caused by benign prostatic hyperplasia (BPH) or prostate cancer. This procedure is typically performed through an incision in the lower abdomen or via a transurethral approach, depending on the patient's condition and the surgeon's preference. The goal is to relieve urinary obstruction and improve urinary flow while minimizing complications. Recovery time varies, but patients often experience significant improvement in urinary symptoms post-surgery. \\ 
nephrectomy & 244 & Nephrectomy is a surgical procedure that involves the removal of a kidney. It can be performed as a partial nephrectomy, where only a portion of the kidney is excised, or as a radical nephrectomy, which entails the complete removal of the kidney along with surrounding tissues, including the adrenal gland and nearby lymph nodes. This procedure is typically indicated for conditions such as kidney cancer, severe kidney damage, or donor kidney retrieval. Nephrectomy can be performed using open surgery or minimally invasive techniques, such as laparoscopic surgery, depending on the patient's condition and the surgeon's expertise. \\ 
partial nephrectomy & 67 & Partial nephrectomy is a surgical procedure that involves the removal of a portion of a kidney while preserving the remaining healthy tissue. This technique is typically employed to treat localized kidney tumors or lesions, allowing for the excision of cancerous or diseased tissue while maintaining kidney function. The procedure can be performed using open surgery or minimally invasive techniques, such as laparoscopic or robotic-assisted surgery. Postoperative care focuses on monitoring kidney function and managing any potential complications. \\ 
inguinal hernia repair & 27 & Inguinal hernia repair is a surgical procedure aimed at correcting an inguinal hernia, which occurs when tissue, often part of the intestine, protrudes through a weak spot in the abdominal muscles in the groin area. The surgery can be performed using an open technique or laparoscopically, where small incisions are made, and a camera is used to guide the repair. The surgeon repositions the protruding tissue back into the abdomen and reinforces the abdominal wall, typically using mesh to provide additional support and reduce the risk of recurrence. Postoperative care includes monitoring for complications and managing pain, with most patients able to return to normal activities within a few weeks. \\ 
nephroureterectomy & 11 & Nephroureterectomy is a surgical procedure that involves the removal of a kidney (nephrectomy) along with the entire ureter, which is the tube that carries urine from the kidney to the bladder. This procedure is typically performed to treat conditions such as kidney cancer, ureteral cancer, or severe urinary tract infections that do not respond to other treatments. The surgery can be done through an open approach or laparoscopically, depending on the patient's condition and the surgeon's expertise. Postoperative care is essential to monitor for complications and ensure proper recovery. \\ 
\bottomrule
\end{tabular}
\vspace{-8pt}
\caption{\textbf{Procedure definitions used to provide clinical context.} GPT\mbox{-}4o was prompted to produce concise, clinically oriented summaries for each procedure. These texts were encoded with the same surgery-pretrained video–language backbone used for the video stream (Sec.~\ref{subsec:clinical_context}).}
\label{tab:proc_definitions}
\end{table*}

\section{Task Definitions Used for Clinical Context}
\label{app:task_defs}
To condition the text encoder with situational priors (Sec.~\ref{subsec:clinical_context}), we expanded task names into concise, clinically phrased summaries using GPT\mbox{-}4o. Tables ~\ref{tab:task_definitions_1}, ~\ref{tab:task_definitions_2}, ~\ref{tab:task_definitions_3}, ~\ref{tab:task_definitions_4} lists the definitions used during inference.

\begin{table*}[t]
\centering
\small
\setlength{\tabcolsep}{8pt}
\renewcommand{\arraystretch}{1.15}
\begin{tabular}{%
    >{\raggedright\arraybackslash}p{0.20\textwidth} 
    >{\raggedright\arraybackslash}p{0.05\textwidth} 
    >{\raggedright\arraybackslash}p{0.72\textwidth} 
}
\toprule
\textbf{Local Task} & \textbf{Count} & \textbf{Definition (GPT-4o-generated)} \\
\midrule
Bladder Neck & 123 & The bladder neck is the region where the bladder connects to the urethra, and during surgical teaching, it is crucial to demonstrate the anatomy and function of this area, emphasizing its role in urinary continence and the potential complications that can arise from surgical manipulation, such as in prostatectomy procedures. Instruct trainees on identifying the bladder neck's landmarks and techniques for preserving its integrity during surgery to minimize postoperative complications. \\
SV & 54 & In urology surgery, the teaching step "SV" refers to "Surgical Visualization," which emphasizes the importance of clear and effective visualization of the surgical field through proper positioning, retraction, and use of appropriate lighting and magnification techniques to enhance the surgeon's ability to identify anatomical structures and perform precise interventions. \\
Posterior Plane & 6 & The posterior plane refers to the anatomical space located behind the bladder and prostate, which is critical for accessing the pelvic structures during urological surgeries. In teaching this step, emphasize the importance of identifying and carefully dissecting the posterior plane to avoid injury to surrounding nerves and vessels while facilitating access to the surgical site. \\
Pedicle/NVB & 24 & The Pedicle/Nerve Vascular Bundle (NVB) step involves careful identification and preservation of the neurovascular structures during urological surgery, particularly in procedures like prostatectomy, to minimize the risk of nerve damage and maintain erectile function. Surgeons should be trained to meticulously dissect around these structures, using appropriate techniques to ensure their integrity while achieving the surgical goals. \\
Apical Dissection & 10 & Apical dissection involves carefully separating the tissue at the apex of the prostate to access the surrounding structures while preserving critical neurovascular bundles. This step is crucial for minimizing complications and ensuring adequate oncological control during prostatectomy procedures. \\
DVC & 7 & DVC, or "Dorsal Venous Complex," refers to the anatomical structure that is critical in urological surgeries, particularly during penile procedures. Teaching this step involves instructing trainees on the identification and management of the DVC to prevent excessive bleeding and ensure proper surgical technique during operations such as circumcision or penile reconstruction. \\
Oblique Checks & 8 & Oblique checks involve assessing the orientation and positioning of the surgical instruments and anatomical structures during a urological procedure to ensure proper alignment and access. This step is crucial for minimizing complications and optimizing the surgical approach. \\
VUA & 154 & Vesicourethral anastomosis (VUA) is the surgical step where the bladder is reconnected to the urethra after prostatectomy or bladder surgery, ensuring proper alignment and tension-free suturing to facilitate optimal healing and urinary function. This step typically involves careful dissection, identification of anatomical landmarks, and the use of absorbable sutures to secure the anastomosis. \\
Adenoma Dissection & 115 & Adenoma dissection involves carefully identifying and separating the adenomatous tissue from surrounding structures using precise surgical techniques, ensuring minimal trauma to adjacent healthy tissue. This step is crucial for achieving complete resection while preserving functional anatomy and reducing the risk of complications. \\
Lymph Node Dissection & 197 & Lymph node dissection involves the surgical removal of lymph nodes to assess for cancer spread, typically performed in conjunction with other procedures such as radical prostatectomy or cystectomy. The teaching step includes identifying the anatomical landmarks, ensuring proper technique to minimize complications, and understanding the implications of lymphatic drainage in urological malignancies. \\

\bottomrule
\end{tabular}
\vspace{-8pt}
\caption{\textbf{(1) Local task definitions used to provide clinical context alongside the procedure.} GPT\mbox{-}4o was prompted to produce concise, clinically oriented summaries for each procedure. These texts were encoded with the same surgery-pretrained video–language backbone used for the video stream (Sec.~\ref{subsec:clinical_context}).}
\label{tab:task_definitions_1}
\end{table*}

\begin{table*}[t]
\centering
\small
\setlength{\tabcolsep}{8pt}
\renewcommand{\arraystretch}{1.15}
\begin{tabular}{%
    >{\raggedright\arraybackslash}p{0.20\textwidth} 
    >{\raggedright\arraybackslash}p{0.05\textwidth} 
    >{\raggedright\arraybackslash}p{0.72\textwidth} 
}
\toprule
\textbf{Local Task} & \textbf{Count} & \textbf{Definition (GPT-4o-generated)} \\
\midrule
Dropping Bladder & 104 & Dropping the bladder involves carefully mobilizing the bladder from its anatomical attachments to the pelvic structures, ensuring that the surrounding tissues are preserved while providing adequate exposure for the surgical procedure. This step is crucial for accessing the underlying structures and minimizing the risk of injury during surgery. \\
Endopelvic Fascia & 32 & The endopelvic fascia is a connective tissue layer that supports pelvic organs and is crucial in urological surgeries. During teaching, emphasize its anatomical significance, the importance of preserving it to maintain pelvic support, and techniques for identifying and dissecting it safely to avoid complications. \\
Pedicles/NVB/Apical Dissection & 5 & During the pedicles, neurovascular bundles (NVB), and apical dissection step, the surgeon carefully identifies and preserves the neurovascular structures while mobilizing the prostate and surrounding tissues. This involves meticulous dissection to minimize damage to the nerves and blood vessels, ensuring optimal functional outcomes post-surgery. \\
Bowel Mobilization & 131 & Bowel mobilization involves carefully dissecting and freeing the bowel from surrounding structures to create a clear surgical field, ensuring that the bowel can be safely retracted or manipulated without compromising its blood supply or function. This step is crucial for accessing the surgical site and minimizing the risk of postoperative complications. \\
Dissection & 181 & Dissection involves carefully separating and isolating anatomical structures using surgical instruments, ensuring minimal trauma to surrounding tissues while providing clear access to the target area for further surgical intervention. This step requires a thorough understanding of anatomy and meticulous technique to avoid complications. \\
Stapling Ureter & 1 & The teaching step for stapling the ureter involves carefully aligning the ureteral edges to ensure proper tension and orientation, followed by the application of a surgical stapler to create a secure and watertight anastomosis. It is crucial to assess the integrity of the staple line and ensure there are no obstructions or kinks in the ureter post-stapling. \\
Bagging Specimen & 4 & Bagging the specimen involves carefully placing the excised tissue or organ into a sterile bag to prevent contamination and facilitate safe handling during transport to pathology. This step is crucial for maintaining specimen integrity and ensuring accurate diagnostic evaluation. \\
Cystotomy & 17 & Cystotomy is the surgical procedure involving an incision into the bladder to access its interior, typically performed to remove bladder stones, tumors, or to facilitate other surgical interventions. During teaching, emphasize the importance of proper incision technique, maintaining hemostasis, and ensuring bladder integrity to prevent complications. \\
Closing Bladder & 112 & Closing the bladder involves carefully suturing the bladder wall using absorbable sutures in a continuous or interrupted fashion, ensuring proper alignment of the tissue layers to prevent leakage and promote healing. It is essential to maintain hemostasis and avoid tension on the sutures to minimize complications. \\
Hemostasis/Bladder Neck & 1 & In the hemostasis step at the bladder neck, the surgeon meticulously identifies and controls any bleeding vessels using techniques such as cauterization or ligation, ensuring a clear surgical field and minimizing the risk of postoperative complications. This step is crucial for maintaining hemostatic balance and facilitating a safe and effective surgical procedure. \\
\bottomrule
\end{tabular}
\vspace{-8pt}
\caption{\textbf{(2) Local task definitions used to provide clinical context alongside the procedure.} GPT\mbox{-}4o was prompted to produce concise, clinically oriented summaries for each procedure. These texts were encoded with the same surgery-pretrained video–language backbone used for the video stream (Sec.~\ref{subsec:clinical_context}).}
\label{tab:task_definitions_2}
\end{table*}

\begin{table*}[t]
\centering
\small
\setlength{\tabcolsep}{8pt}
\renewcommand{\arraystretch}{1.15}
\begin{tabular}{%
    >{\raggedright\arraybackslash}p{0.20\textwidth} 
    >{\raggedright\arraybackslash}p{0.05\textwidth} 
    >{\raggedright\arraybackslash}p{0.72\textwidth} 
}
\toprule
\textbf{Local Task} & \textbf{Count} & \textbf{Definition (GPT-4o-generated)} \\
\midrule

DVC, Hemostasis, Rocco Stitch & 5 & DVC (Dorsal Venous Complex) management involves careful identification and control of the dorsal venous complex to minimize bleeding during prostatectomy. Hemostasis is achieved by meticulously ligating or cauterizing blood vessels to prevent excessive blood loss, while the Rocco stitch is a surgical technique used to reinforce the posterior urethral support by suturing the bladder neck to the pelvic floor, enhancing continence postoperatively. \\
DVC, Hemostasis & 1 & DVC (Dorsal Venous Complex) hemostasis involves identifying and controlling the dorsal venous complex during urological procedures, typically by clamping, ligating, or cauterizing the vessels to prevent excessive bleeding. This step is crucial for maintaining a clear surgical field and ensuring patient safety throughout the operation. \\
Bladder Neck (Trigonization?) & 7 & Bladder neck trigonization involves the surgical technique of reshaping or reconstructing the bladder neck to improve urinary function, often performed during procedures for conditions like bladder outlet obstruction or incontinence. The teaching step emphasizes the importance of identifying anatomical landmarks, ensuring proper tension on the bladder neck, and achieving a secure closure to maintain urinary continence postoperatively. \\
Tumor Dissection & 1 & Tumor dissection involves carefully separating the tumor from surrounding tissues and structures while preserving critical anatomical landmarks and minimizing damage to adjacent organs. This step requires meticulous technique to ensure complete tumor removal and to reduce the risk of recurrence or complications. \\
Suturing Kidney & 1 & When suturing the kidney, begin by selecting an appropriate suture material and needle, then carefully approximate the renal tissue edges using a continuous or interrupted technique, ensuring to avoid excessive tension to preserve blood supply and function. Finally, secure the sutures with adequate knots, ensuring hemostasis and proper alignment of the renal parenchyma. \\
Peritoneal Flap and Mesh Pocket Dissection & 11 & Peritoneal flap and mesh pocket dissection involves carefully mobilizing the peritoneum to create a flap that can be used to cover a surgical site, while simultaneously dissecting a pocket for the placement of mesh. This step requires meticulous attention to avoid injury to surrounding structures and ensure adequate space for secure mesh placement, promoting optimal healing and minimizing complications. \\
Anchoring of Mesh & 4 & Anchoring of mesh involves securely attaching the surgical mesh to the surrounding tissue to provide support and prevent migration, typically using sutures or fixation devices. This step is crucial for ensuring the stability and effectiveness of the mesh in reinforcing the pelvic floor or repairing hernias. \\
Closing Peritoneum & 12 & Closing the peritoneum involves carefully approximating the edges of the peritoneal layer using absorbable sutures, ensuring that the tissue is aligned properly to promote healing and minimize the risk of adhesions. It is important to maintain a tension-free closure while avoiding excessive manipulation of surrounding structures. \\
Bladder Closure & 7 & Bladder closure involves meticulously suturing the bladder wall to ensure a secure and watertight seal, typically using absorbable sutures in a continuous or interrupted fashion, while carefully avoiding tension on the tissue to promote optimal healing and prevent complications such as leakage or stricture. It is essential to assess the integrity of the closure before concluding the procedure, often by performing a saline test to check for any leaks. \\

\bottomrule
\end{tabular}
\vspace{-8pt}
\caption{\textbf{(3) Local task definitions used to provide clinical context alongside the procedure.} GPT\mbox{-}4o was prompted to produce concise, clinically oriented summaries for each procedure. These texts were encoded with the same surgery-pretrained video–language backbone used for the video stream (Sec.~\ref{subsec:clinical_context}).}
\label{tab:task_definitions_3}
\end{table*}

\begin{table*}[t]
\centering
\small
\setlength{\tabcolsep}{8pt}
\renewcommand{\arraystretch}{1.15}
\begin{tabular}{%
    >{\raggedright\arraybackslash}p{0.20\textwidth} 
    >{\raggedright\arraybackslash}p{0.05\textwidth} 
    >{\raggedright\arraybackslash}p{0.72\textwidth} 
}
\toprule
\textbf{Local Task} & \textbf{Count} & \textbf{Definition (GPT-4o-generated)} \\
\midrule

SVs & 50 & SVs, or surgical videos, are utilized as a teaching step to provide visual demonstrations of surgical techniques and procedures. They serve as an educational tool to enhance understanding of anatomy, instrumentation, and surgical maneuvers, allowing trainees to visualize complex steps before performing them in the operating room. \\
SV/Vas & 20 & The SV/Vas teaching step involves the identification and dissection of the seminal vesicles (SV) and vas deferens (Vas) during urological surgery, ensuring careful preservation of surrounding structures while facilitating access for procedures such as vasectomy or prostatectomy. This step emphasizes the importance of anatomical landmarks and the technique of blunt and sharp dissection to minimize complications. \\
Bladder Neck (Suture) & 4 & The bladder neck suture step involves placing sutures at the bladder neck to provide support and maintain proper anatomical alignment, which is crucial for preventing postoperative complications such as urinary incontinence. This step requires careful technique to ensure the sutures are secure yet not overly tight, allowing for normal bladder function. \\
NVB & 7 & NVB, or Neurovascular Bundle, refers to the critical anatomical structures that include nerves and blood vessels surrounding the prostate. In surgical teaching, it is essential to identify and preserve the NVB during prostatectomy to maintain erectile function and minimize complications, emphasizing careful dissection and awareness of anatomical landmarks. \\
Pexy & 7 & Pexy refers to the surgical technique of anchoring or securing an organ or tissue to another structure, often used in procedures like cystopexy or nephropexy to stabilize the bladder or kidney, respectively. In teaching this step, emphasize the importance of proper anatomical identification, tension management, and the choice of suturing technique to ensure effective stabilization while minimizing complications. \\
Defatting Prostate & 2 & Defatting the prostate involves carefully removing excess adipose tissue surrounding the prostate gland to enhance visibility and access during surgery. This step is crucial for minimizing bleeding and facilitating the subsequent surgical procedures, such as prostatectomy or other interventions. \\
Leak Testing, etc & 2 & Leak testing involves filling the surgical site or reconstructed structure with a fluid (usually saline) to assess for any leaks or defects in the anastomosis or closure. This step is crucial to ensure the integrity of the surgical repair before finalizing the procedure, as it helps identify potential complications that could lead to postoperative issues. \\
\bottomrule
\end{tabular}
\vspace{-8pt}
\caption{\textbf{(4) Local task definitions used to provide clinical context alongside the procedure.} GPT\mbox{-}4o was prompted to produce concise, clinically oriented summaries for each procedure. These texts were encoded with the same surgery-pretrained video–language backbone used for the video stream (Sec.~\ref{subsec:clinical_context}).}
\label{tab:task_definitions_4}
\end{table*}



\section{Ontology: Mapping Cluster Tags to Individual Mentions}
\label{app:ontology_tag_to_mention}

\subsection{Instrument Ontology: Cluster--to--Mention Mapping}
\label{app:instrument_mapping}
Table~\ref{tab:instrument_cluster_mapping} lists the mapping between \emph{instrument} cluster tags and the raw surface forms (with counts) mined from trainer$\rightarrow$trainee feedback. These mappings were used to normalize lexical variation during ontology construction (Sec.~\ref{sec:ontology_from_raw_feedback}).

\begin{table*}[t]
\small
\centering
\setlength{\tabcolsep}{6pt}
\renewcommand{\arraystretch}{1.15}
\begin{tabular}{%
  >{\raggedright\arraybackslash}p{0.22\textwidth}%
  >{\raggedright\arraybackslash}p{0.74\textwidth}%
}
\toprule
\textbf{Instrument cluster tag} & \textbf{Individual mentions (count)} \\
\midrule
\texttt{energy\_device} &
\texttt{electrocautery (59); hook (6); energy device (3); bovie (2); coagulation tool (1); clip (1); coagulator (1); monopolar (1); bipolar (1); vessel sealer (1); bovy (1); buzz (1); energy devices (1)} \\
\addlinespace[2pt]
\texttt{fourth\_arm} &
\texttt{fourth arm (32); 4th arm (30); stitch (2); retractor (1); scissors (1)} \\
\addlinespace[2pt]
\texttt{left\_hand} &
\texttt{left hand (86); left arm (2); right hand (2); scissors (2); 4th arm (1); fourth arm (1); left hand instrument (1)} \\
\addlinespace[2pt]
\texttt{needle} &
\texttt{needle (33); needles (3); balloon (1); suture (1)} \\
\addlinespace[2pt]
\texttt{stitch} &
\texttt{stitch (29); stitches (9); chromic stitch (1); foley (1)} \\
\addlinespace[2pt]
\texttt{suture\_material} &
\texttt{suture (18); sutures (5); vicryl (1); 3-0 suture (1); vicyl stitch (1)} \\
\bottomrule
\end{tabular}
\caption{\textbf{Instrument cluster tags and their constituent mentions.} Counts in parentheses reflect frequency in the raw feedback corpus; items are shown verbatim (including capitalization and spelling variants).}
\label{tab:instrument_cluster_mapping}
\end{table*}

\subsection{Tissue Ontology: Cluster--to--Mention Mapping}
\label{app:tissue_mapping}
Table~\ref{tab:tissue_cluster_mapping} lists the mapping between \emph{tissue} cluster tags and the raw surface forms (with counts) mined from trainer$\rightarrow$trainee feedback. These mappings were used to normalize lexical variation during ontology construction (Sec.~\ref{sec:ontology_from_raw_feedback}).

\begin{table*}[t]
\small
\centering
\setlength{\tabcolsep}{6pt}
\renewcommand{\arraystretch}{1.15}
\begin{tabular}{%
  >{\raggedright\arraybackslash}p{0.22\textwidth}%
  >{\raggedright\arraybackslash}p{0.74\textwidth}%
}
\toprule
\textbf{Tissue cluster tag} & \textbf{Individual mentions (count)} \\
\midrule
\texttt{adrenal\_kidney\_region} &
\texttt{kidney (10); liver (5); psoas (5); adrenal (2); lateral edge of adrenal (2); cyst (2); adrenal vessels (1); lower pole (1); fascia of psoas (1); psoas sheets (1); pancreas (1); adrenal gland (1); pelvic brim (1); fat plane (1); adrenal area (1); tendon (1); spleen (1)} \\
\addlinespace[2pt]
\texttt{bladder} &
\texttt{bladder (37); bladder neck (13); posterior bladder neck (3); anterior bladder neck (2); prostate (2); bladder fibers (1); central suture (1); vessels (1); abdominal fat (1); needle catheter (1)} \\
\addlinespace[2pt]
\texttt{general\_vasculature} &
\texttt{vessel (27); vessels (10); veins (3); aorta (3); arteries (2); vein (2); pedicles (1); blood vessel (1); lateral structure (1); arterial pumper (1); blood clot (1); epigastric vessel (1); tissue (1); nerves (1)} \\
\addlinespace[2pt]
\texttt{gi\_tract} &
\texttt{mucosa (22); bowel (6); rectum (4); diverticuli (1); mucosal edge (1); colon (1); rectal integrity (1); inner mucosa (1); bowel wall (1); ileostomy (1); ileum (1)} \\
\addlinespace[2pt]
\texttt{major\_veins} &
\texttt{vein (55); veins (4); tissue (2); lumbar vein (1); cava (1); puboprostatic ligaments (1); external iliac vein (1); inferior vena cava (1)} \\
\addlinespace[2pt]
\texttt{muscle\_capsule} &
\texttt{muscle (8); capsule (6); detrusor fibers (3); capsule fibers (2); muscle fiber (1); tendon (1); detrusor muscle (1); muscle fibers (1); rectourethralis muscle (1); diaphragm (1); detrusor (1)} \\
\addlinespace[2pt]
\texttt{prostate\_tissue} &
\texttt{prostate (67); adenoma (11); prostate contour (6); tumor (2); median lobe (2); fat (1); apex (1); peripheral zone (1); vessel (1); tissue (1); small prostate (1); detrusor fibers (1)} \\
\addlinespace[2pt]
\texttt{seminal\_vesicle\_vas} &
\texttt{Seminal Vesicles (18); vas deferens (15); tissue (3); Vas deferens (2); septum (1); VAS (1); spermatic cord (1); medial aspect (1); vessels (1); vesicles (1); medial surface (1)} \\
\addlinespace[2pt]
\texttt{ureter} &
\texttt{ureter (38); gondal (1); gonadal (1); colon (1); fat (1); urether (1)} \\
\addlinespace[2pt]
\texttt{urethra} &
\texttt{urethra (28); bladder (4); sphincter (3); lateral structure (1); posterior plane (1); tissue around urethra (1); periurethral tissue (1); bladder neck (1); urethra edge (1); posterior structure (1); apex (1)} \\
\bottomrule
\end{tabular}
\caption{\textbf{Tissue cluster tags and their constituent mentions.} Counts in parentheses reflect frequency in the raw feedback corpus; items are shown verbatim (including capitalization and spelling variants).}
\label{tab:tissue_cluster_mapping}
\end{table*}

\subsection{Action Ontology: Cluster--to--Mention Mapping}
\label{app:action_mapping}
Tables~\ref{tab:action_cluster_mapping_p1} and \ref{tab:action_cluster_mapping_p2} list the mapping between \emph{action} cluster tags and the raw surface forms (with counts) mined from trainer$\rightarrow$trainee feedback. These mappings were used to normalize lexical variation during ontology construction (Sec.~\ref{sec:ontology_from_raw_feedback}).

\begin{table*}[t]
\small
\centering
\setlength{\tabcolsep}{6pt}
\renewcommand{\arraystretch}{1.15}
\begin{tabular}{%
  >{\raggedright\arraybackslash}p{0.19\textwidth}%
  >{\raggedright\arraybackslash}p{0.77\textwidth}%
}
\toprule
\textbf{Action cluster tag} & \textbf{Individual mentions (count)} \\
\midrule
\texttt{adjust} &
\texttt{get right (3); slow down (2); go slow (2); narrow down (2); match up (1); take (1); increase (1); go slower (1); go (1); ease (1); use less (1); give yourself enough length (1); spend a little extra time (1); narrow it down (1); set up (1); be slower (1); see (1); T it (1)} \\
\addlinespace[2pt]
\texttt{adjust\_direction} &
\texttt{go lateral (4); be more midline (1); Aim back up (1); err on the lateral sides (1); come from the lateral side (1); go obliquely (1); avoid (1); come obliquely (1); come from the sides (1); go cephalad more (1); come across (1); go to the other side (1); go more anterior (1); proceed lateral (1); "go to other side" (1); go anterior (1); get to midline (1); go medial (1); bring right (1); push medially (1); get to the medial (1); stay medial (1); perpendicular (1); start laterally (1); get out laterally (1); march medially (1); cock (1); "go to the other side" (1); extend right (1); come proximal (1); come from the other side (1); march up laterally (1); come to the other side (1); go in (1); work towards midline (1); up (1)} \\
\addlinespace[2pt]
\texttt{apply\_traction} &
\texttt{grab (9); don't pull (3); pull down (3); pull (3); pull back (2); hold (2); take (2); Press (2); stop (2); pull up (2); press (2); get better (1); pull with two hands (1); go (1); pull through (1); pull towards you (1); put (1); take out (1); regrab (1); bring (1); keep pulling down (1); pick up (1); press up (1); drop (1)} \\
\addlinespace[2pt]
\texttt{coagulate} &
\texttt{buzz (56); sweep (2); Buzz (2); don't buzz blindly (1); be patient (1); burn (1); coag (1); buzz on that crease (1); give a little buzz (1); bovie (1); wait (1); buzz directly (1); start buzzing (1); buzz off (1); buzz little (1); do buzzing (1); lift (1); buzz it (1); buzz right above (1); identify (1); pre buzz (1); keep buzzing (1); Buzz higher (1); buzz slowly (1); get around (1); look (1); take (1); go down (1)} \\
\addlinespace[2pt]
\texttt{continue\_action} &
\texttt{give (7); do (5); try (3); use (3); exhaust (1); create more (1); do more (1); use it (1); Do (1); do again (1); practice (1); do one (1); try doing (1); Be proactive (1)} \\
\addlinespace[2pt]
\texttt{control\_depth} &
\texttt{go down (4); go higher (3); come down (2); go deeper (2); keep going down (2); stay up (2); get up (2); take higher (2); try to get (1); come above (1); go lower (1); put one below first (1); stay low (1); work (1); go further down (1); stay above (1); start higher (1); go below (1); get below (1); take (1); go down more (1); go above (1); go right below (1); get lower (1); higher up (1); don't go deep (1); take lower (1); get deeper (1); keep it low (1); stay down (1); get down (1); get under (1); get underneath (1); go deep (1)} \\
\addlinespace[2pt]
\texttt{cut} &
\texttt{cut (15); cut down (3); cold cut (2); do one thing (1); cut just above (1); slice (1); suture cut (1); cut carefully (1); keep cutting (1); start cutting (1); cut through (1)} \\
\addlinespace[2pt]
\texttt{dissect} &
\texttt{get off (7); deliver (2); free up (2); take out (2); take away (1); define (1); bloc it (1); stay sharp (1); go down (1); scrape (1); bag (1); dont scrape (1); get more out (1); get off first (1)} \\
\addlinespace[2pt]
\texttt{ensure\_safety} &
\texttt{be careful (24); rip (5); be gentle (3); don't cut (3); rip carefully (2); be aware (1); dont rip (1); get first (1); err (1); DONT rip (1); rip out (1); spare (1); hit (1); dont be aggressive (1); come right (1); see (1); watch out (1); vigilant (1); rip off (1)} \\
\bottomrule
\end{tabular}
\caption{\textbf{Action cluster tags and their constituent mentions (part 1).} Counts in parentheses reflect frequency in the raw feedback corpus; items are shown verbatim (including capitalization and contractions).}
\label{tab:action_cluster_mapping_p1}
\end{table*}

\begin{table*}[t]
\small
\centering
\setlength{\tabcolsep}{6pt}
\renewcommand{\arraystretch}{1.15}
\begin{tabular}{%
  >{\raggedright\arraybackslash}p{0.24\textwidth}%
  >{\raggedright\arraybackslash}p{0.72\textwidth}%
}
\toprule
\textbf{Action cluster tag} & \textbf{Individual mentions (count)} \\
\midrule
\addlinespace[2pt]
\texttt{grasp} &
\texttt{grab (51); take (44); regrab (5); take it (4); hold (2); hang on (2); take less (2); don't grab (2); look up (2); start (1); take first (1); push up (1); take one more (1); let go (1); grab lower (1); grab over (1); let it fall (1); "Take" (1); get more tissue (1); grab with left hand (1); grab less (1)} \\
\addlinespace[2pt]
\texttt{inspect} &
\texttt{see (10); watch (6); do (2); look (2); reduce (1); see below (1); see posterior plane (1); extract (1); look down (1); grab (1); look higher (1); take (1); watch videos (1); look back (1); do one side (1); see more structures (1); UA back (1); make sure (1)} \\
\addlinespace[2pt]
\texttt{navigate\_space} &
\texttt{come (5); travel more (4); come back (3); get out (2); back up (2); come out (2); bring (2); go back (2); travel (2); travel a little bit (2); come back a little bit (1); inch a little bit (1); get right on (1); come in a little bit (1); travel here (1); come back a little more (1); proceed up (1); come more (1); don't travel so much (1); move (1); Rock out a little bit (1); travel a little bit more (1); come up (1); go more (1); bring back up (1); go back down (1); bring in (1); enter back (1); travel less (1); rock back (1); move back (1); come back down (1); get in (1); get away (1); get the tip out (1); get on the plane (1); go in (1)} \\
\addlinespace[2pt]
\texttt{open} &
\texttt{open (45); open up (8); open more (3); open up a little bit more (2); open up wide (1); show (1); open completely (1); Get tips in (1); make (1); forget (1); keep going (1); let go (1); open out a little bit more (1); stop (1); open laterally (1); open spaces (1); open wide (1); come out (1); extend (1)} \\
\addlinespace[2pt]
\texttt{perform\_sweeping\_maneuvers} &
\texttt{sweep (23); sweep down (5); sweep off (3); don't sweep (2); sweep up (2); sweep towards you (2); swipe (2); sweep back (2); preserve (1); swipe up (1); Sweep it down (1); sweep further (1); gently sweep (1); sweep away (1); stay (1); sweep more (1); drive (1); sweep towards (1); sweep again (1); spread transversely (1); lateral sweep (1); be gentle (1); bluntly sweep off (1); do shorter sweeps (1); gently sweep back (1); sweep shorter (1); start from top down (1); sweep laterally (1); use more energy (1)} \\
\addlinespace[2pt]
\texttt{release\_instrument} &
\texttt{drop (11); leave (6); leave it (2); drop down (1); leave it be (1); leave a little bit (1); leave alone (1); leave it alone (1); Release a little bit (1); leave a little more (1); finish dropping (1); start (1); get in (1)} \\
\addlinespace[2pt]
\texttt{rotate} &
\texttt{rotate (6); reverse (1); Rotate (1); do (1)} \\
\addlinespace[2pt]
\texttt{spread} &
\texttt{separate (4); widen (3); make it wide (1); extend distally (1); skinny down (1); thin out (1); don't go too wide (1); open (1); open up (1); go right (1); go wide (1); spread (1); put (1); let go (1); don't go wide (1)} \\
\addlinespace[2pt]
\texttt{stop} &
\texttt{stop (48); stop short (1); Stop (1)} \\
\addlinespace[2pt]
\texttt{suture} &
\texttt{throw (5); sew (1); don't pull (1); loop up (1); take stitch (1); bigger loop (1); throw sutures (1); take (1); loop (1)} \\
\addlinespace[2pt]
\texttt{turn} &
\texttt{turn (19); flip (5); flip up (2); U it back (1); Turn more (1); curl (1); go in (1); round (1); Grab (1); turn more (1); try turning (1); turn towards (1); turn up (1); learn (1); curve (1); turn out (1)} \\

\bottomrule
\end{tabular}
\caption{\textbf{Action cluster tags and their constituent mentions (part 2).} Counts in parentheses reflect frequency in the raw feedback corpus; items are shown verbatim (including capitalization and contractions).}
\label{tab:action_cluster_mapping_p2}
\end{table*}





\end{document}